# Massively Parallel Universal Linear Transformations using a Wavelength-Multiplexed Diffractive Optical Network

Jingxi Li[1,2,3], Bijie Bai[1,2,3], Yi Luo[1,2,3], and Aydogan Ozcan[1,2,3*]

[1]Electrical and Computer Engineering Department, University of California, Los Angeles, CA, 90095, USA

[2]Bioengineering Department, University of California, Los Angeles, CA, 90095, USA

[3]California NanoSystems Institute (CNSI), University of California, Los Angeles, CA, 90095, USA

[*]Correspondence to: ozcan@ucla.edu

## Abstract

Large-scale linear operations are the cornerstone for performing complex computational tasks. Using optical computing to perform linear transformations offers potential advantages in terms of speed, parallelism and scalability. Previously, the design of successive spatially-engineered diffractive surfaces forming an optical network was demonstrated to perform statistical inference and compute an arbitrary complex-valued linear transformation using narrowband illumination. Here, we report deep learning-based design of a massively parallel broadband diffractive neural network for all-optically performing a large group of arbitrarily-selected, complex-valued linear transformations between an input and output field-of-view, each with $N_i$ and $N_o$ pixels, respectively. This broadband diffractive processor is composed of $N_w$ wavelength channels, each of which is uniquely assigned to a distinct target transformation; a large set of arbitrarily-selected linear transformations can be individually performed through the same diffractive network at different illumination wavelengths, either simultaneously or sequentially (wavelength scanning). We demonstrate that such a broadband diffractive network, regardless of its material dispersion, can successfully approximate $N_w$ unique complex-valued linear transforms with a negligible error when the number of diffractive neurons ($N$) in its design is $\geq 2N_w N_i N_o$. We further report that the spectral multiplexing capability ($N_w$) can be increased by increasing $N$; our numerical analyses confirm these conclusions for $N_w > 180$, which can be further increased to e.g., ~2000 depending on the upper bound of the approximation error. Massively parallel, wavelength-multiplexed diffractive networks will be useful for designing high-throughput intelligent machine vision systems and hyperspectral processors that can perform statistical inference and analyze objects/scenes with unique spectral properties.

# Introduction

Computing plays an increasingly vital role in constructing intelligent, digital societies. The exponentially growing power consumption of digital computers brings some important challenges for large-scale computing. Optical computing can potentially provide advantages in terms of power efficiency, processing speed and parallelism. Motivated by these, we have witnessed various research and development efforts on advancing optical computing over the last few decades[1–32]. Synergies between optics and machine learning have enabled the design of novel optical components using deep learning-based optimization[33–44], while also allowing the development of advanced optical/photonic information processing platforms for artificial intelligence[5,20–32,45].

Among different optical computing designs, diffractive optical neural networks represent a free-space-based framework that can be used to perform computation, statistical inference, and inverse design of optical elements[22]. A diffractive neural network is composed of multiple transmissive and/or reflective diffractive layers (or surfaces), which leverage light-matter interactions to jointly perform modulation of the input light field to generate the desired output field. These passive diffractive layers, each containing thousands of spatially-engineered diffractive features (termed as "diffractive neurons"), are designed (optimized) in a computer using deep learning tools, e.g., stochastic gradient descent and error back-propagation. Once the training process converges, the resulting diffractive layers are fabricated to form a passive, free-space optical processing unit that does not consume any power except the illumination light. This framework is also scalable since it can adapt to changes in the input field-of-view (FOV) or data dimensions by adjusting the size and/or the number of diffractive layers. Diffractive networks can directly access the 2D/3D input information of a scene or object, and process the optical information encoded in, e.g., the amplitude, phase, spectrum and polarization of the input light, making them highly suitable as intelligent optical front-ends for machine vision systems.

Diffractive neural networks have been used to perform various optical information processing tasks, including, e.g., object classification[22,46–57], image reconstruction[52,58,59], all-optical phase recovery and quantitative image phase imaging[60], class-specific imaging[61], super-resolution image displays[62], and logical operations[63–65]. Employing successive spatially-engineered diffractive surfaces as the backbone for inverse design of deterministic optical elements also enabled numerous applications such as spatially-controlled wavelength demultiplexing[66], pulse engineering[67], and orbital angular momentum multiplexing/demultiplexing[68].

In addition to these task-specific applications, diffractive networks also serve as general-purpose computing modules that can be used to create compact, power-efficient all-optical processors. Recent efforts have shown that a diffractive network can be used to all-optically perform an arbitrarily-selected, complex-valued linear transformation between its input and output FOV with a negligible error when the number of trainable diffractive neurons ($N$) approaches $N_i N_o$, where $N_i$ and $N_o$ represent the number of pixels at the input and output FOVs, respectively[69]. By using non-trainable, predetermined polarizer arrays within an isotropic diffractive network, a polarization-encoded diffractive processor was also demonstrated to accurately perform a group of $N_p = 4$ distinct complex-valued linear transformations using a single system with $N \geq N_p N_i N_o = 4 N_i N_o$; in this case, each one of these 4 optical transformations can be accessed through a different combination of the input/output polarization states[70]. This polarization-encoded diffractive system is limited to a multiplexing factor of $N_p = 4$ since an additional desired transformation matrix that can be assigned to a new combination of input-output polarization states can be written as a linear combination of the four linear transforms that are already learned by the diffractive processor[70]. These former works involved *monochromatic* diffractive networks where a single illumination wavelength encoded the input information channels.

Here, we report the use of a wavelength multiplexing scheme to create a broadband diffractive optical processor, which massively increases the throughput of all-optical computing by performing a group of distinct linear transformations in parallel using a single diffractive network. By encoding the input/output information of the target linear transforms using $N_w$ different wavelengths (i.e., $\lambda_1, \lambda_2, \dots, \lambda_{N_w}$), we created a single broadband diffractive network to simultaneously perform a group of $N_w$ arbitrarily-selected, complex-valued linear transforms with a negligible error. We demonstrate that $N \geq 2N_w N_i N_o$ diffractive neurons are required to successfully implement $N_w$ complex-valued linear transforms using a broadband diffractive processor, where the thickness values of its diffractive neurons constitute the *only* variables optimized during the deep learning-based training process. Without loss of generality, we numerically demonstrate wavelength-division multiplexed (WDM) universal linear transformations with $N_w > 180$, which can be further increased to ~2000 based on the approximation error threshold that is acceptable. We also demonstrate that these wavelength-multiplexed universal linear transformations can be implemented even with a flat material dispersion, where the refractive index (n) of the material at the selected wavelength channels is the same, i.e., $n(\lambda) \approx n_o$ for $\lambda \in \{\lambda_1, \lambda_2, \dots, \lambda_{N_w}\}$. The training process of these WDM diffractive networks was adaptively balanced across different wavelengths of operation such that the all-optical linear transformation accuracies of the different channels were similar to each other, without introducing a bias towards any wavelength channel or the corresponding linear transform.

It is important to emphasize that the goal of this work is not to train the broadband diffractive network to implement the correct linear transformations for only a few input-output field pairs. We are *not* aiming to use the diffractive layers as a form of metamaterial that can output different images or optical fields at different wavelengths. Instead, our goal is to generalize the performance of our broadband diffractive processor to *infinitely many* pairs of input and output complex fields that satisfy the target linear transformation at each spectral channel, thus achieving *universal all-optical computing* of multiple complex-valued matrix-vector multiplications, accessed by a set of illumination wavelengths ($N_w \gg 1$).

Compared to previous literature, this manuscript has various unique aspects: (1) this is the first demonstration of a spatially-engineered diffractive system to achieve spectrally-multiplexed universal linear transformations; (2) the level of massive multiplexing that is reported through a single WDM diffractive network (e.g., $N_w > 180$) is significantly larger compared to other channels of multiplexing, including e.g., polarization diversity[70], and this number can be further increased to e.g., $N_w \approx 2000$ with more diffractive neurons ($N$) used in the network design; (3) deep learning-based training of the diffractive layers used adaptive spectral weights to equalize the performances of all the linear transformations assigned to $N_w$ different wavelengths; (4) the capability to perform multiple linear transformations using wavelength multiplexing does not require any wavelength-sensitive optical elements to be added into the diffractive network design, except for wavelength scanning or broadband illumination with demultiplexing filters; and (5) this wavelength-multiplexed diffractive processor can be implemented using various materials with different dispersion properties (including materials with a flat dispersion curve), and is widely applicable to different parts of the electromagnetic spectrum including the visible band.

Based on the massive parallelism exhibited by this broadband diffractive network, we believe that this platform and the underlying concepts can be used to develop optical processors operating at different parts of the spectrum with extremely high computing throughput. Its throughput can be further increased by expanding the range and/or the number of encoding wavelengths as well as combining WDM with other multiplexing schemes such as polarization encoding. The reported framework would be valuable for the development of multi-color and hyperspectral machine vision systems that perform statistical inference based on the spatial and spectral information of an object or a scene, which may find applications in various fields, including e.g., biomedical imaging, remote sensing, analytical chemistry, and material science.

## Results

### Design of wavelength-multiplexed diffractive optical networks for massively parallel universal linear transformations

Throughout this manuscript, the terms "diffractive deep neural network", "diffractive neural network", "diffractive optical network" and "diffractive network" are interchangeably used. Figure 1 illustrates the schematic of our broadband diffractive optical network design for massively parallel, wavelength-multiplexed all-optical computing. The broadband diffractive network, composed of 8 successive diffractive layers, contains in total $N$ diffractive neurons with their thickness values as learnable variables, which are jointly trained to perform a group of $N_w$ linear transformations between the input and output FOVs through $N_w$ parallel wavelength channels. More details about this diffractive architecture, its optical forward model and training details can be found in the Methods section. To start with, a group of $N_w$ different wavelengths, $\lambda_1, \lambda_2, \ldots, \lambda_{N_w}$, are selected to be used as the wavelength channels for the broadband diffractive processor to encode different input complex fields and perform different target transformations; see Fig. 1. For the implementation of the broadband diffractive designs in this paper, we fixed the mean value $\lambda_m$ of this group of wavelengths $\{\lambda_1, \lambda_2, \ldots, \lambda_{N_w}\}$, i.e., $\lambda_m = \frac{1}{N_w}\sum_{w=1}^{N_w}\lambda_w$, and assigned these wavelengths to be equally spaced between $\lambda_1 = 0.9125\lambda_m$ and $\lambda_{N_w} = 1.0875\lambda_m$. Unless otherwise specified, we chose $\lambda_m$ to be 0.8 mm in our numerical simulations, as it aligns with the terahertz band that was experimentally used in several of our previous work[50,52,58,59,61,62,66,67]. Without loss of generality, the wavelengths used for the design of the broadband diffractive processors can also be selected at other parts of the electromagnetic spectrum, such as the visible band, for which the related simulation results and analyses can be found in the Discussion section to follow. Based on the scalar diffraction theory, the broadband optical fields propagating in the diffractive system are simulated at these selected wavelengths using a sampling period of $0.5\lambda_m$ along both the horizontal and vertical directions. We also select $0.5\lambda_m$ as the size of the individual neurons on the diffractive layers. With these selections, we include in our optical forward model all the propagating modes that are transmitted through the diffractive layers.

Let $\boldsymbol{i}$ and $\boldsymbol{o}'$ be the complex-valued, vectorized versions of the 2D input and output broadband complex fields at the input and output FOVs of the diffractive network, respectively, as shown in Fig. 1. We denote $\boldsymbol{i}_w$ and $\boldsymbol{o}'_w$ as the complex fields generated by sampling the optical fields at the wavelength $\lambda_w$ ($w \in \{1, 2, \ldots, N_w\}$) within the input and output FOVs, respectively, and then vectorizing the resulting 2D matrices in column-major order. According to this notation, $\boldsymbol{i}_w$ and $\boldsymbol{o}'_w$ represent the input and output of the $w^{\text{th}}$ wavelength channel in our WDM diffractive network, respectively. In the following analyses, without loss of generality, the number of pixels at the input and output FOVs are selected to be the same, i.e., $N_i = N_o$.

To implement $N_w$ target linear transformations, we randomly generated $N_w$ complex-valued matrices $A_1, A_2, ..., A_{N_w}$, each composed of $N_i \times N_o$ entries, to serve as a group of unique arbitrary linear transformations to be all-optically implemented using a wavelength-multiplexed diffractive processor. All these matrices, $A_1, A_2, ..., A_{N_w}$, are generated using unique random seeds to ensure that they are different; we further confirmed the differences between these randomly generated matrices by calculating the cosine similarity values between any two combinations of the matrices in a given set (see e.g., Supplementary Fig. S1). For each unique matrix $A_w \in \{A_1, A_2, ..., A_{N_w}\}$, we randomly generated a total of 70,000 complex-valued input field vectors $\{i_w\}$, and created the corresponding output field vectors $\{o_w\}$ by calculating $o_w = A_w i_w$. We separated these input-output complex field pairs into three individual sets for training, validation and testing, each containing 55,000, 5,000 and 10,000 samples, respectively. By increasing the size of these training datasets to e.g., >100,000 input-output pairs of randomly generated complex fields, it is possible to further improve the transformation accuracy of the trained broadband diffractive networks; since this does not change the general conclusions of this work, it is left as future work. More details on the generation of the training and testing data can be found in the Methods section.

Based on the notations introduced above, the objective of training our wavelength-multiplexed diffractive processor is that, for any of its wavelength channels operating at $\lambda_w$ ($w \in \{1, 2, .., N_w\}$), the diffractive output fields $\{o'_w\}$ computed from any given inputs $\{i_w\}$ should provide a match to the output ground truth (target) fields $\{o_w\}$. If this can be achieved for any arbitrary choice of $\{i_w\}$, this means that the all-optical transformations $A'_w$ performed by the trained broadband diffractive system at different wavelength channels constitute an accurate approximation to their ground truth (target) transformation matrices $A_w$ where $w \in \{1, 2, .., N_w\}$.

As the first step of our analysis, we selected the input/output field size to be $N_i = N_o = 8 \times 8 = 64$, and started to train broadband diffractive processors with $N_w = 2, 4, 8, 16$ and 32 wavelength channels. Results and analysis of implementing more wavelength channels (*e.g.*, $N_w > 100$) through a single diffractive processor will be provided in later subsections. For this task, we randomly generated a set of 32 different matrices with dimensions of $64 \times 64$, i.e., $A_1, A_2, ..., A_{32}$, with their first eight visualized (as examples) in Fig. 2a with their amplitude and phase components. Supplementary Fig. S1a also reports the cosine similarity values between these randomly generated 32 matrices confirming that they are all very close to 0. For each $N_w$ mentioned above, we also trained several broadband diffractive designs with different numbers of trainable diffractive neurons, i.e., $N \in \{3.9k, 8.2k, 16.9k, 32.8k, 64.8k, 131.1k, 265.0k\}$, all using the same training datasets $\{(i_w, o_w)\}$, randomly generated based on the target transformations $\{A_w\}$ ($w \in \{1, 2, ..., N_w\}$), and the same number of training epochs.

To benchmark the performances of these wavelength-multiplexed diffractive networks, for each $N$ we also trained *monochromatic* diffractive networks without using any wavelength multiplexing as our baseline, which can approximate only one target linear transformation using a single wavelength (i.e., $N_w = 1$). Here, we simply select $\lambda_m$ as the operating wavelength of this baseline monochrome diffractive network used for comparison.

During the training of these diffractive networks, mean-squared error (MSE) loss is calculated per wavelength channel to make the diffractive output fields come as close to the ground truth (target) fields as possible. However, in the wavelength-multiplexed diffractive models, treating all these channels equally in the final loss function would result in the all-optical transformation accuracies being biased since longer wavelengths present lower spatial resolution. To address this issue and equalize the all-optical transformation accuracies of all the wavelengths within the selected channel set, we devised a strategy by adaptively adjusting the weight coefficients applied to the loss terms of these channels during the training process (see the Methods section for details).

After the deep learning-based training of the broadband diffractive designs introduced above is completed, the resulting all-optical diffractive transformations of these models are summarized in Fig. 2b-d. We quantified the generalization performance of these broadband diffractive networks on the blind testing dataset for each transformation using three different metrics: (1) the normalized transformation MSE ($MSE_{\text{Transformation}}$), (2) the cosine similarity ($CosSim$) between the all-optical transforms and the target transforms, and (3) the mean-squared error between the diffractive network output fields and their ground truth output fields ($MSE_{\text{Output}}$)[53,69]. More details about the definitions of these performance metrics are provided in the Methods section. For the diffractive designs with different numbers of wavelength channels ($N_w$ = 1, 2, 4, 8, 16 and 32), we report these performance metrics in Fig. 2b-d as a function of the number of trainable diffractive neurons ($N$). These performance metrics reported in Fig. 2 refer to the mean values calculated across all the wavelength channels, while the results of the individual wavelength channels are shown in Fig. 3.

In Fig. 2b, it can be seen that the transformation errors of all the trained diffractive models show a monotonic decrease as $N$ increases, which is expected due to the increased degrees of freedom in the diffractive processor. Also, the approximation errors of the regular diffractive networks without using wavelength multiplexing, i.e., $N_w = 1$, approaches 0 as $N$ approaches $2N_iN_o \approx 8.2$k. This observation confirms the conclusion obtained in our previous work[69,70], i.e., a phase-only monochrome diffractive network requires at least $2N_iN_o$ diffractive neurons to approximate a target complex-valued linear transformation with a negligible error. On the other hand, for the wavelength-multiplexed diffractive models with $N_w$ different wavelength channels that are trained to approximate $N_w$ unique linear transforms, we see in Fig. 2 that the approximation errors approach 0 as $N$ approaches $2N_wN_iN_o$. This finding indicates that compared to a baseline monochrome diffractive model that can only perform a *single* transform, performing *multiple* distinct transforms using wavelength multiplexing within a single diffractive network requires its number of trainable neurons $N$ to be increased by $N_w$-fold. This conclusion is further supported by the results of the other two performance metrics, $CosSim$ and $MSE_{\text{Output}}$, as shown in Fig. 2c and d: as $N$ approaches $2N_wN_iN_o$, $CosSim$ and $MSE_{\text{Output}}$ of the wavelength-multiplexed diffractive models approach 1 and 0, respectively.

To reveal the linear transformation performance of the individual wavelength channels in our wavelength-multiplexed diffractive processors, in Fig. 3 we show the channel-wise output field errors ($MSE_{\text{Output}}$) of the wavelength-multiplexed diffractive networks with $N_w$ = 2, 4, 8, 16 and 32 and $N = 2N_wN_iN_o$. Figure 3 indicates that the $MSE_{\text{Output}}$ of these individual channels are very close to each other in all the designs with different $N_w$, demonstrating no significant performance bias toward any specific wavelength channel or target transform. For comparison, we also show in Supplementary Fig. S2 the resulting $MSE_{\text{Output}}$ of the diffractive model with $N_w = 8$ and $N = 2N_wN_iN_o = 16N_iN_o$ when our channel balancing training strategy with adaptive weights was *not* used (see the Method section). There appears to be a large variation at the output field errors among the different wavelength channels if adaptive weights were not used during the training; in fact, the channels assigned to longer wavelengths tend to show much inferior transformation performance, which highlights the significance of using our balancing strategy during the training process. Stated differently, unless a channel balancing strategy is employed during the training phase, longer wavelengths suffer from relatively lower spatial resolution and face increased all-optical transformation errors compared to the shorter wavelength channels.

To visually demonstrate the success of our broadband diffractive system in performing a group of linear transformations using wavelength multiplexing, in Fig. 4 we show examples of the ground truth transformation matrices (i.e., $A_w$) and their all-optical counterparts (i.e., $A'_w$) resulting from the diffractive designs with $N_w = 8$ and $N \in \{2N_w N_i N_o = 32 N_i N_o = 64.8\text{k}; 4N_w N_i N_o = 16 N_i N_o = 131.1\text{k}\}$. The amplitude and phase absolute errors between the two ($A_w$ and $A'_w$) are also reported in the same figure. Moreover, in Fig. 5 and Supplementary Fig. S3 we present some exemplary complex-valued input-output optical fields from the same set of diffractive designs with $N = 4N_w N_i N_o = 131.1\text{k}$ and $N = 2N_w N_i N_o = 64.8\text{k}$, respectively. These results, summarized in Figs. 4-5 and Supplementary Fig. S3, reveal that, when $N \geq 2N_w N_i N_o$, the all-optical transformation matrices and the output complex fields of all the wavelength channels match their ground truth targets very well with a negligible error, which is also in line with our earlier observations in Fig. 2.

**Limits of $N_w$: Scalability of wavelength-multiplexing in diffractive networks**

We have so far demonstrated that a single broadband diffractive network can be designed to simultaneously perform a group of $N_w$ arbitrary complex-valued linear transformations, with $N_w = 2, 4, 8, 16$ and $32$ (Figs. 2 and 3). Next, we explore the feasibility of implementing a significantly larger number of wavelength channels in our system to better understand the limits of $N_w$. Due to our limited computational resources available, to simulate the behavior of larger $N_w$ values we selected $N_i = N_o = 5 \times 5$ and $N_w \in \{1, 2, 4, 8, 16, 32, 64, 128 \text{ and } 184\}$. Accordingly we generated a new set of 184 different arbitrarily selected complex-valued matrices with dimensions of $25 \times 25$, i.e., $A_1, A_2, \ldots, A_{184}$, as the target linear transformations to be all-optically implemented. The cosine similarity values between these randomly generated matrices are reported in Supplementary Fig. S1b, confirming that they are all very close to 0. We also created training, validation and testing datasets based on these new target transformation matrices following the same approach as in the previous subsection: for each transformation matrix we randomly generated 55,000, 5,000 and 10,000 field samples for the training, validation and testing datasets, respectively. Then, using the training field datasets, we trained broadband diffractive designs with $N_w$ different wavelength channels, where the $N_w$ target transforms were taken from the first $N_w$ matrices in the randomly generated set $\{A_1, A_2, \ldots, A_{184}\}$. For each $N_w$ choice, we also trained diffractive models with different number of diffractive neurons, including $N = 1.5 N_w N_i N_o$, $N = 2 N_w N_i N_o$ and $N = 3 N_w N_i N_o$.

The all-optical transformation performance metrics of the resulting diffractive networks on the testing datasets are shown in Fig. 6 as a function of $N_w$. Figs. 6a-c reveal that the all-optical transformations of the diffractive designs with different $N$ show some increased error as $N_w$ increases. For the diffractive models with $N = 3 N_w N_i N_o$, the all-optical transformation errors ($MSE_{\text{Transformation}}$) at smaller $N_w$ appear to be extremely small, and do not exhibit the same performance degradation with increasing $N_w$; only after $N_w > 10$ we see an error increase in the all-optical transformations for $N = 3 N_w N_i N_o$. By comparing the linear transformation performance of the models with different $N$, Fig. 6 clearly reveals that adding more diffractive neurons to a broadband diffractive network design can greatly improve its transformation performance, which is especially important to operate at a large $N_w$.

By having a linear fit to the data points shown in Figs. 6a and c, we can extrapolate to larger $N_w$ values and predict an all-optical transformation error bound as a function of $N_w$. With these fitted (dashed) lines shown in Figs. 6a and c, we get a coarse prediction of the linear transformation performance of a broadband diffractive model with a significantly larger number of wavelength channels $N_w$ that is challenging to simulate due to our limited computer memory and speed. Interestingly, these three fitted lines (corresponding to diffractive designs with $N = 1.5 N_w N_i N_o$, $N = 2 N_w N_i N_o$ and $N = 3 N_w N_i N_o$) intersect with each other at a point around $N_w = 10{,}000$ with an $MSE_{\text{Transformation}}$ of ~0.2 and an $MSE_{\text{Output}}$ of ~0.03. This level of transformation error coincides with the error levels observed at the beginning of our training, implying that a broadband diffractive model with $N_w = {\sim}10{,}000$, even after training, would only exhibit a performance level comparable to an untrained model. These analyses indicate that, for a broadband diffractive network trained with $N \leq 3 N_w N_i N_o$ and a training dataset of 55K optical field pairs, there is an empirical multiplexing upper bound of $N_w = {\sim}10{,}000$.

However, before reaching this $N_w = {\sim}10{,}000$ ultimate limit discussed above, practically the desired level of approximation accuracy will set the actual limit of $N_w$. For example, based on visual inspection and the calculated PSNR (peak signal-to-noise ratio) values, one can empirically choose a blind testing error of $MSE_{\text{Output}} {\sim} 10^{-3}$ as a *threshold* for the diffractive network's all-optical approximation error; this threshold corresponds to a mean PSNR value of ~20 dB, calculated for the diffractive network output fields against their ground truth (see Supplementary Fig. S4). We marked this $MSE_{\text{Output}}$-based performance threshold in Fig. 6c using a black dashed line, which also corresponds to a transformation error ($MSE_{\text{Transformation}}$) of ~$9\times10^{-3}$, which was also marked in Fig. 6a with a black dashed line. Based on these empirical performance thresholds set by $MSE_{\text{Output}} \approx 10^{-3}$ and PSNR $\approx$ 20 dB, we can infer that a broadband diffractive processor with $N = 3 N_w N_i N_o$ can accommodate up to $N_w {\sim} 2{,}000$ wavelength channels, where ~2,000 different linear transformations can be performed through a single broadband diffractive processor within the performance bounds shown in Figs. 6a and c; see the purple dashed lines. The same analysis reveals a reduced upper bound of $N_w {\sim} 600$ for the diffractive network designs with $N = 2 N_w N_i N_o$; see the green dashed lines.

**Impact of material dispersion and losses on wavelength-multiplexed diffractive networks**

In the previous subsection, we showed that a broadband diffractive processor can be designed to implement >180 different target linear transforms simultaneously, and this number can be further extended to e.g., $N_w {\sim} 2{,}000$ based on an all-optical approximation error threshold of $MSE_{\text{Output}} \approx 10^{-3}$. In this subsection, we provide additional analyses on material-related factors that have an impact on the accuracy of wavelength-multiplexed computing through broadband diffractive networks. For example, the selection of materials with different dispersion properties (i.e., the real and imaginary parts of the refractive index as a function of the wavelength) will impact the light-matter interactions at different illumination wavelengths. To numerically explore the impact of material dispersion and related optical losses, we took the broadband diffractive network design shown in Fig. 6 with $N_w = 128$ and $N = 3 N_w N_i N_o$, and re-trained it using different materials. The first material we selected is a lossy polymer, which is widely employed as a 3D printing material; this material was used to fabricate diffractive networks that operate at the THz part of the spectrum[52,66,67]. The dispersion curves of this lossy material are shown in Supplementary Fig. S5a, which were also used in the design of the diffractive networks reported in the previous sub-sections (with $\lambda_m = 0.8$ mm). As a second material choice for comparison, we selected a lossless dielectric material, for which we took N-BK7 glass as an example and used its dispersion to simulate our wavelength-multiplexed diffractive processor design at the visible wavelengths with $\lambda_m = 530$ nm; the dispersion curves of this material are reported in Supplementary Fig. S5b. As a third material choice for comparison, we considered a hypothetical scenario where the material of the diffractive layers had a flat dispersion at around $\lambda_m = 0.8$ mm, with no absorption and a constant refractive index (~1.72) across all the selected wavelength channels of interest; see the refractive index curve of this "dispersion-free" material in Supplementary Fig. S5c.

After the training of the diffractive network models using these different materials selected for comparison, we summarized their all-optical linear transformation performance in Fig. 7a-c; see the purple bars. These results reveal that all the three diffractive models with different material choices achieved negligible all-optical transformation errors, regardless of their dispersion characteristics. This confirms the feasibility of extending our wavelength-multiplexed diffractive processor designs to other spectral bands with vastly different material dispersion features.

In addition to the all-optical transformation accuracy, the output diffractive efficiency ($\eta$) of these diffractive network models is also practically important. As shown in Fig. 7d, due to the absorption by the layers, the diffractive network model using the lossy polymer material presents a very poor output diffraction efficiency, $\eta$, compared to the other two diffractive models that used lossless materials. In addition to the absorption of light through the diffractive layers, a WDM diffractive network also suffers from optical losses due to the propagating waves that leak out of the diffractive processor volume. This second source of optical loss within a diffractive network can be strongly mitigated through the incorporation of diffraction efficiency-related penalty terms[52,66,67,69] into the training loss function (see the Methods for details). The results of using such a diffraction-efficiency related penalty term during training are presented in Fig. 7a-d (yellow bars), which indicate that the output diffraction efficiencies of the corresponding models were improved by e.g., >589-1479 fold compared to their counterparts that were trained without using such a penalty term (see Fig. 7d). We also show in Fig. 7e and 7f the output diffraction efficiencies of the individual wavelength channels trained without and with the diffraction efficiency penalty term, respectively. These results also revealed that the diffraction efficiency-related penalty term used during training not only improved the overall output efficiency of the diffractive processor design, but also helped to mitigate the imbalance of diffraction efficiencies among different wavelength channels (see Fig. 7e-f). These improvements also come at an expense; as shown in Fig. 7a-c, there is some degradation in the all-optical transformation performance of the diffractive networks that are trained with a diffraction efficiency-related penalty term. However, this relative degradation in the all-optical transformation performance is still acceptable since e.g., a cosine similarity value of >0.996-0.998 is maintained in each case (see Fig. 7b, yellow bars).

**Impact of limited bit-depth on the accuracy of wavelength-multiplexed diffractive networks**

The bit depth of a broadband diffractive network refers to the finite number of thickness levels that each diffractive neuron can have on top of a common base thickness of each diffractive layer. For example, in a broadband diffractive network with a bit depth of 8, its diffractive neurons will be trained to have at most $2^8 = 256$ different thickness values that are distributed between a predetermined minimum thickness and a maximum thickness value. To mechanically support each diffractive layer, the minimum thickness is always positive, acting as the base thickness of each layer. To analyze the impact of this bit depth on the linear transformation performance and accuracy of our wavelength-multiplexed diffractive networks, we took the $N_w$=184 channel diffractive design reported in the previous subsections (trained using a data format with 32-bit-depth), and re-trained it from scratch under different bit depths, including 4, 8, 12. Based on the same test dataset, the all-optical linear transformation performance metrics of the resulting diffractive networks are reported in Fig. 8 as a function of $N$. Figure 8 reveals that a 12 bit depth is practically identical to using a 32 bit depth in terms of the all-optical transformation accuracy that can be achieved for the $N_w$=184 target linear transformations. Furthermore, a bit depth of 8 can also be used for a broadband diffractive network design to maintain its all-optical transformation performance with a relatively small error increase, which can be compensated for with an increase in $N$ as illustrated in Fig. 8. These observations from Fig. 8 highlight (1) the importance of having a sufficient bit depth in the design and fabrication of a broadband diffractive processor; (2) the importance of $N$ as a way to boost the all-optical transformation performance under a limited diffractive neuron bit depth.

**Impact of wavelength precision or jitter on the accuracy of wavelength-multiplexed diffractive networks**

Another possible factor that may cause systematic errors in our framework is the wavelength precision or jitter. To analyze the wavelength encoding related errors, we used the 4-channel wavelength-multiplexed diffractive network model with $N \approx 2N_w N_i N_o = 8N_i N_o$ and $N_i = N_o = 8^2$ that was presented in Fig. 3b. We deliberately shifted the illumination wavelength used for each encoding channel away from the pre-selected wavelength used during the training (i.e., $\lambda_1 = 0.9125 \lambda_m$, $\lambda_2 = 0.9708 \lambda_m$, $\lambda_3 = 1.0292 \lambda_m$, and $\lambda_4 = 1.0875 \lambda_m$). The resulting linear transformation performance of the $N_w = 4$ channels using different performance metrics are summarized in Figs. 9a-c as a function of the illumination wavelength. All of these results in Fig. 9 show that as the illumination wavelengths used for each encoding channel gradually deviate from their designed/assigned wavelengths (used during the training of the WDM diffractive network), their all-optical transformation accuracy begins to degrade. To shed more light on this, we used the previous performance threshold based on $MSE_{\text{Output}} \approx 10^{-3}$ as an empirical criterion to estimate the tolerable range of illumination wavelength errors, which revealed an acceptable bandwidth of ~$0.002\lambda_m$ for each one of the encoding wavelength channels. Stated differently, when a given illumination wavelength is within ± ~$0.001\lambda_m$ of the corresponding pre-selected wavelength assigned for that spectral channel, the degradation of the linear transformation accuracy at the output of the WDM diffractive network will satisfy $MSE_{\text{Output}} \leq 10^{-3}$. In practical applications, this level of spectral precision can be routinely achieved by using high-performance wavelength scanning sources[71,72] (e.g., swept-source lasers) or narrow passband thin-film filters.

**Discussion**

We demonstrated wavelength-multiplexed diffractive network designs that can perform massively parallel universal linear transformations through a single diffractive processor. We also quantified the limits of $N_w$ and the impact of material dispersion, bit depth and wavelength precision/jitter on the all-optical transformation performance of broadband diffractive networks. In addition to these, other factors may limit the performance of broadband diffractive processors, including the lateral and axial misalignments of diffractive layers, surface reflections, and other imperfections introduced during the fabrication. To mitigate some of these practical issues, various approaches such as high-precision lithography and anti-reflection coatings can be utilized in the fabrication process of a diffractive network. As demonstrated in our previous work[50,52,61], it is also possible to mitigate the performance degradation resulting from some of these experimental factors by incorporating them as random errors into the physical forward model used during the training process, which is referred to as "vaccination" of the diffractive network.

To the best of our knowledge, there has not been a demonstration of a design for the all-optical implementation of a complex-valued, arbitrary linear transformation using metasurfaces or metamaterials. In principle, having different diffractive meta-units placed on the same substrate to perform different transformations at different wavelengths could be attempted as an alternative approach to what we presented in this manuscript. However, such an approach would face severe challenges since (1) at large spectral multiplexing factors ($N_w \gg 1$) shown in this work, the lateral period for each spectral meta-design will substantially increase per substrate, lowering the accuracy of each transformation; (2) at each illumination wavelength, the other meta-units designed for (assigned to) the other spectral components, will also introduce "cross-talk fields" that will severely contaminate the desired responses in each wavelength and cannot be neglected since $N_w \gg 1$; (3) the phase responses of the spectrally-encoded meta-units, in general, cover a small angular range, leading to low numerical aperture (NA) solutions compared to the diffractive solutions reported in this work, where NA = 1 (in air); the low NA of meta-units fundamentally limits the space-bandwidth product of each transformation channel; (4) if multiple layers of metasurfaces are used in a given design, all of these aforementioned sources of errors associated with spectral meta-units will accumulate and get amplified through the subsequent field propagation in a cascaded manner, causing severe degradations to the final output fields, compared to the desired fields. Perhaps due to these significant challenges outlined here, metasurface or metamaterial-based diffractive designs have not yet been reported as a solution to perform universal linear transformations - neither an arbitrary complex-valued linear transformation nor a group of linear transformations through some form of multiplexing.

As we have shown in the Results section, a diffractive neuron number of $N \geq 2N_w N_i N_o$ is required for a WDM diffractive network to successfully implement $N_w$ different complex-valued linear transforms. Compared to the previous complex-valued monochrome ($N_w = 1$) diffractive designs[69], the additional factor of 2 in $N$ results from the fact that the only trainable degrees of freedom for a broadband WDM diffractive design are the thickness values of the diffractive neurons, while the $N_w$ different target transformations are all complex-valued. Stated differently, the resulting modulation values of different wavelengths through each diffractive neuron are mutually coupled through the dispersion of the material and depend on the neuron thickness.

Finally, we would like to emphasize that this presented framework can operate at various parts of the electromagnetic spectrum, including the visible band, so that the set of wavelength channels used to perform the transformation multiplexing can match with the light source and/or the spectral signals emitted from or reflected by the objects. In practice, this massively parallel linear transformation capability can be utilized in an optical processor to perform distinct statistical inference tasks using different wavelength channels, bringing in additional throughput and parallelism to optical computing. This highly multiplexed WDM diffractive network design might also inspire the development of new multi-color and hyperspectral machine vision systems for the identification, encoding, and reconstruction of various objects that have unique spectral properties, with numerous applications in e.g., biomedical imaging, remote sensing, analytical chemistry, and material science.

## Methods

**Forward model of the broadband diffractive neural network.** A wavelength-multiplexed diffractive network consists of successive diffractive layers that collectively modulate the incoming broadband optical fields. In the forward model of our numerical simulations, the diffractive layers are assumed to be thin optical modulation elements, where the $m^{\text{th}}$ feature on the $k^{\text{th}}$ layer at a spatial location $(x_m, y_m, z_m)$ represents a wavelength-dependent complex-valued transmission coefficient, $t^k$, given by:

$$t^k(x_m, y_m, z_m, \lambda) = a^k(x_m, y_m, z_m, \lambda)\exp\left(j\phi^k(x_m, y_m, z_m, \lambda)\right) \quad (1),$$

where $a$ and $\phi$ denote the amplitude and phase coefficients, respectively. The diffractive layers are connected to each other by free-space propagation, which is modeled through the Rayleigh-Sommerfeld diffraction equation[22,46]:

$$f_m^k(x, y, z) = \frac{z - z_i}{r^2}\left(\frac{1}{2\pi r} + \frac{1}{j\lambda}\right)\exp\left(\frac{j2\pi r}{\lambda}\right) \quad (2),$$

where $f_m^k(x, y, z, \lambda)$ is the complex-valued field on the $m^{\text{th}}$ pixel of the $k^{\text{th}}$ layer at $(x, y, z)$ at a wavelength of $\lambda$, which can be viewed as a secondary wave generated from the source at $(x_m, y_m, z_m)$; and $r = \sqrt{(x - x_m)^2 + (y - y_m)^2 + (z - z_m)^2}$ and $j = \sqrt{-1}$. For the $k^{\text{th}}$ layer ($k \geq 1$, assuming the input plane is the $0^{\text{th}}$ layer), the modulated optical field $E^k$ at location $(x_m, y_m, z_m)$ is given by:

$$E^k(x_m, y_m, z_m, \lambda) = t^k(x_m, y_m, z_m) \cdot \sum_{n \in S} E^{k-1}(x_n, y_n, z_n, \lambda) \cdot f_m^{k-1}(x_m, y_m, z_m) \quad (3),$$

where $S$ denotes all the diffractive neurons on the previous diffractive layer.

We chose $\lambda_m/2$ as the smallest sampling period for the simulation of the complex optical fields, and also used $\lambda_m/2$ as the smallest feature size of the diffractive layers. In the input and output FOVs, a $4 \times 4$ binning is performed on the simulated optical fields, resulting in a pixel size of $2\lambda_m$ for the input/output fields. The axial distance ($d$) between the successive layers (including the diffractive layers and the input/output planes) in our diffractive processor designs is empirically selected as $d = 0.5D_{\text{Layer}}$, where $D_{\text{Layer}}$ represents the lateral size of each diffractive layer.

The diffractive thickness value, $h$, of each neuron of a diffractive layer is composed of two parts, $h_{\text{learnable}}$ and $h_{\text{base}}$, as follows:

$$h = h_{\text{learnable}} + h_{\text{base}} \quad (4),$$

where $h_{\text{learnable}}$ denotes the learnable thickness parameters of each diffractive feature and is confined between $h_{\text{min}} = 0$ and $h_{\text{max}} = 1.25\lambda_m$. When a modulation with q-bit depth is applied to the diffractive model, $h_{\text{learnable}}$ will be rounded to the nearest number that corresponds to one of $2^q$ different equally-spaced levels within the range of $[0, h_{\text{learnable}}]$. The additional base thickness, $h_{\text{base}}$, is a constant, which is chosen as $0.25\lambda_{\text{m}}$ to serve as substrate support for the diffractive neurons. To achieve the constraint applied to $h_{\text{learnable}}$, an associated latent trainable variable $h_v$ was defined using the following analytical form:

$$h_{\text{learnable}} = \frac{h_{\max}}{2} \cdot (\sin(h_v) + 1) \tag{5}$$

Note that before the training starts, $h_v$ values of all the diffractive neurons were randomly initialized with a normal distribution (a mean value of 0 and a standard deviation of 1). Based on these definitions, the amplitude and phase components of the complex transmittance of $m^{\text{th}}$ feature of layer $l$, i.e., $a^k(x_m, y_m, z_m, \lambda)$ and $\phi^k(x_m, y_m, z_m, \lambda)$, can be written as a function of the thickness of each neuron $h_m$ and the incident wavelength $\lambda$:

$$a^k(x_m, y_m, z_m, \lambda) = \exp\left(-\frac{2\pi\kappa(\lambda)h_m^k}{\lambda}\right) \tag{6}$$

$$\phi^k(x_m, y_m, z_m, \lambda) = (n(\lambda) - n_{\text{air}})\frac{2\pi h_m^k}{\lambda} \tag{7}$$

where the wavelength-dependent parameters $n(\lambda)$ and $\kappa(\lambda)$ are the refractive index and the extinction coefficient of the diffractive layer material corresponding to the real and imaginary parts of the complex-valued refractive index $\tilde{n}(\lambda)$, i.e., $\tilde{n}(\lambda) = n(\lambda) + j\kappa(\lambda)$ [66]. In this work, we considered three different materials to form the diffractive layers of a broadband diffractive processor, including a lossy polymer, a lossless dielectric and a hypothetical lossless dispersion-free material. Among these, the lossy polymer material represents a UV-curable 3D printing material (VeroBlackPlus RGD875, Stratasys Ltd.), which was used in our previous work[52,66,67] for 3D-printing of diffractive networks. The lossless dielectric material, used for the diffractive models operating at the visible band, represents N-BK7 glass (Schott), ignoring the negligible absorption through thin layers. The dispersion-free material, on the other hand, assumed a lossless material with its refractive index $n(\lambda)$ having a flat distribution with respect to the wavelength range of interest, i.e., $n(\lambda) \approx 1.72$. The final $n(\lambda)$ and $\kappa(\lambda)$ curves of different materials that were used for training the diffractive models reported in this paper are shown in Supplementary Fig. S5.

**Preparation of the linear transformation datasets.** In this paper, the input and output FOVs of the diffractive networks are assumed to have the same size, which is set as $8 \times 8$ or $5 \times 5$ pixels based on the assigned linear transformation tasks, i.e., $\boldsymbol{i}_w, \boldsymbol{o}_w \in \mathbb{C}^{8\times 8}$ or $\mathbb{C}^{5\times 5}$ ($w \in \{1, 2, \ldots, N_w\}$). Accordingly, the size of the target complex-valued transformation matrices $\boldsymbol{A}_w$ is equal to $64 \times 64$ or $25 \times 25$, respectively, i.e., $\boldsymbol{A}_w \in \mathbb{C}^{64\times 64}$ ($w \in \{1, 2, \ldots, 32\}$) or $\boldsymbol{A}_w \in \mathbb{C}^{25\times 25}$ ($w \in \{1, 2, \ldots, 184\}$). The amplitude and phase components of all these target matrices $\boldsymbol{A}_w$ were generated with a uniform ($U$) distribution of $U[0, 1]$ and $U[0, 2\pi]$, respectively, using the pseudo-random number generation function *random.uniform()* built-in NumPy. Different random seeds were used to generate these transformation matrices to ensure they were uniquely different. For training a broadband diffractive network with $N_w$ wavelength channels, the amplitude and phase components of the input fields $\boldsymbol{i}_w$ ($w \in \{1, 2, \ldots, N_w\}$) were randomly generated with a uniform ($U$) distribution of $U[0, 1]$ and $U[0, 2\pi]$, respectively. The ground truth (target) fields $\boldsymbol{o}_w$ ($w \in$ $\{1, 2, \dots, N_w\}$) were generated by calculating $\boldsymbol{o}_w = \boldsymbol{A}_w \boldsymbol{i}_w$. For each $\boldsymbol{A}_w$ ($w \in \{1, 2, \dots, N_w\}$), we generated a total of 70,000 input/output complex optical fields to form a dataset, which was then divided into three parts: training, validation, and testing, each containing 55,000, 5,000, and 10,000 complex-valued optical field pairs, respectively.

**Training loss function.** For each wavelength channel, the normalized MSE loss function is defined as:

$$\mathcal{L}_{\text{MSE},w} = E\left[\frac{1}{N_o} \sum_{n=1}^{N_o} |\widehat{\boldsymbol{o}_w}[n] - \widehat{\boldsymbol{o}'_w}[n]|^2\right]$$

$$= E\left[\frac{1}{N_o} \sum_{n=1}^{N_o} |\sigma_w \boldsymbol{o}_w[n] - \sigma'_w \boldsymbol{o}'_w[n]|^2\right] \quad (8),$$

where $E[\cdot]$ denotes the average across the current batch, $w$ stands for the $w^{\text{th}}$ wavelength channel that is being accessed, and $[n]$ indexes the $n^{\text{th}}$ element of the vector. $\sigma_w$ and $\sigma'_w$ are the coefficients used to normalize the energy of the ground truth (target) field $\boldsymbol{o}_w$ and the diffractive-network output field $\boldsymbol{o}'_w$, respectively, which are given by:

$$\sigma_w = \frac{1}{\sqrt{\sum_{n=1}^{N_o} |\boldsymbol{o}_w[n]|^2}} \quad (9),$$

$$\sigma'_w = \frac{\sum_{n=1}^{N_o} \sigma_w \boldsymbol{o}_w[n] \boldsymbol{o}'^*_w[n]}{\sum_{n=1}^{N_o} |\boldsymbol{o}'_w[n]|^2} \quad (10).$$

During the training of each broadband diffractive network, all the wavelength channels are simultaneously simulated, and the training data are fed into the channels at the same time. The WDM diffractive network is trained based on the loss averaged across different wavelength channels. The total loss function $\mathcal{L}$ that we used can be written as:

$$\mathcal{L} = \frac{1}{N_w} \sum_{w=1}^{N_w} \alpha_w \mathcal{L}_{\text{MSE},w} \quad (11),$$

where $\alpha_w$ is the adaptive spectral weight coefficient applied to the loss for the $w^{\text{th}}$ wavelength channel, which was used to balance the performances achieved by different wavelength channels during the optimization process. The initial values of $\alpha_w$ for all the wavelength channels are set as 1. After the optimization begins, $\alpha_w$ is adaptively updated after each training step using the following formula:

$$\alpha_w \leftarrow \max(0.1 \times (\mathcal{L}_{\text{MSE},w} - \mathcal{L}_{\text{MSE},w_{\text{ref}}}) + \alpha_w, 0) \quad (12),$$

where $\mathcal{L}_{\text{MSE},w_{\text{ref}}}$ represents the MSE loss of the wavelength channel that is chosen to be a *reference* to measure the difference in the loss of the other channels. This also means that $\alpha_w$ for the wavelength channel selected as the *reference* remains unchanged at 1. For the trained broadband diffractive models presented in this paper, we chose the middle channel as the reference wavelength channel, i.e., $w_{\text{ref}} = N_w/2$. According to this approach, for a wavelength channel $w$ that is not a reference channel, when the loss of the channel is small compared to that of the reference channel, $\alpha_w$ will automatically decrease to reduce the weight of the corresponding channel. Conversely, when the loss of a specific wavelength channel is large compared to that of the reference channel, $\alpha_w$ will automatically grow to increase the weight of the channel and thus enhance the subsequent penalty on the corresponding channel performance.

In order to increase the output diffraction efficiencies of the diffractive networks, we incorporated an additional efficiency penalty term to the loss function of Eq. 11:

$$\mathcal{L} = \frac{1}{N_w} \sum_{w=1}^{N_w} (\alpha_w \mathcal{L}_{\text{MSE},w} + \beta \mathcal{L}_{\text{Eff},w}) \tag{13},$$

where $\mathcal{L}_{\text{Eff},w}$ represents the diffraction efficiency penalty loss applied to the $w^{\text{th}}$ wavelength channel, and $\beta$ represents its weight, empirically set as $10^4$. $\mathcal{L}_{\text{Eff},w}$ is defined as:

$$\mathcal{L}_{\text{Eff},w} = \begin{cases} \eta_{\text{th}} - \eta_w, & \text{if } \eta_{\text{th}} \geq \eta_w \\ 0, & \text{if } \eta_{\text{th}} < \eta_w \end{cases} \tag{14},$$

where $\eta_w$ represents the mean output diffraction efficiency for the $w^{\text{th}}$ wavelength channel of the WDM diffractive network, and $\eta_{\text{th}}$ refers to a predetermined penalization threshold, which was taken as $3\times10^{-5}$ (for diffractive models using the lossy polymer materials) or $3\times10^{-4}$ (for the other diffractive models using lossless dielectric or dispersion-free materials). $\eta_w$ is defined as:

$$\eta_w = E\left[\frac{\sum_{n=1}^{N_o}|o'_w[n]|^2}{\sum_{n=1}^{N_i}|i_w[n]|^2}\right] \tag{15}.$$

**Performance metrics used for the quantification of the all-optical transformation errors.** To quantitatively evaluate the transformation results of the wavelength-multiplexed diffractive networks, four different performance metrics were calculated per wavelength channel of the diffractive designs using the blind testing data set: (1) the normalized transformation mean-squared error ($MSE_{\text{Transformation}}$), (2) the cosine similarity ($CosSim$) between the all-optical transforms and the target transforms, (3) the normalized mean-squared error between the diffractive network output fields and their ground truth ($MSE_{\text{Output}}$), and (4) the output diffraction efficiency (Eq. 15). The transformation error for the $w^{\text{th}}$ wavelength channel of the WDM diffractive network, $MSE_{\text{Transformation},w}$, is defined as:

$$\begin{aligned} MSE_{\text{Transformation},w} &= \frac{1}{N_i N_o} \sum_{n=1}^{N_i N_o} |\boldsymbol{a}_w[n] - m_w \boldsymbol{a}'_w[n]|^2 \\ &= \frac{1}{N_i N_o} \sum_{n=1}^{N_i N_o} \left|\boldsymbol{a}_w[n] - \widehat{\boldsymbol{a}'_w}[n]\right|^2 \end{aligned} \tag{16},$$

where $\boldsymbol{a}_w$ is the vectorized version of the ground truth (target) transformation matrix assigned to the $w^{\text{th}}$ wavelength channel $\boldsymbol{A}_w$, i.e., $\boldsymbol{a}_w = \text{vec}(\boldsymbol{A}_w)$. $\boldsymbol{a}'_w$ is the vectorized version of $\boldsymbol{A}'_w$, which is the all-optical transformation matrix performed by the trained diffractive network. $m_w$ is a scalar coefficient used to eliminate the effect of diffraction efficiency-related scaling mismatch between $\boldsymbol{A}_w$ and $\boldsymbol{A}'_w$, i.e.,

$$m_w = \frac{\sum_{n=1}^{N_i N_o} \boldsymbol{a}_w[n] \boldsymbol{a}'^*_w[n]}{\sum_{n=1}^{N_i N_o} |\boldsymbol{a}'_w[n]|^2} \tag{17}.$$

The cosine similarity between the all-optical diffractive transform and its target (ground truth) for the $w^{\text{th}}$ wavelength channel, $CosSim_w$, is defined as:

$$CosSim_w = \frac{|\boldsymbol{a}_w^H \widehat{\boldsymbol{a}}'_w|}{\sqrt{\sum_{n=1}^{N_i N_o} |\boldsymbol{a}_w[n]|^2} \sqrt{\sum_{n=1}^{N_i N_o} |\widehat{\boldsymbol{a}'_w}[n]|^2}} \tag{18}.$$

The normalized mean-squared error between the diffractive network outputs and their ground truth for the $w^{\text{th}}$ wavelength channel, $MSE_{\text{Output},w}$, is defined using the same formula as in Eq. 8, except that $E[\cdot]$ is calculated across the entire testing set.

**Training-related details.** All the diffractive optical networks used in this work were trained using PyTorch (v1.11.0, Meta Platforms Inc.). We selected AdamW optimizer[73,74] for training all the models, and its parameters were taken as the default values and kept identical in each model. The batch size was set as 8. The learning rate, starting from an initial value of 0.001, was set to decay at a rate of 0.5 every 10 epochs, respectively. The training of the diffractive network models was performed with 50 epochs. The best models were selected based on the MSE loss calculated on the validation data set. For the training of our diffractive models, we used a workstation with a GeForce RTX 3090 graphical processing unit (GPU, Nvidia Inc.) and Intel® Core™ i9-12900F central processing unit (CPU, Intel Inc.) and 64 GB of RAM, running Windows 11 operating system (Microsoft Inc.). The typical time required for training a wavelength-multiplexed diffractive network model with e.g., $N_w = 128$ and $N = 1.5 N_w N_i N_o$ is ~50 hours.

# References

1  Solli DR, Jalali B. Analog optical computing. *Nat Photonics* 2015; **9**: 704–706.

2  Wetzstein G, Ozcan A, Gigan S, Fan S, Englund D, Soljačić M et al. Inference in artificial intelligence with deep optics and photonics. *Nature* 2020; **588**: 39–47.

3  Shastri BJ, Tait AN, Ferreira de Lima T, Pernice WHP, Bhaskaran H, Wright CD et al. Photonics for artificial intelligence and neuromorphic computing. *Nat Photonics* 2021; **15**: 102–114.

4  Zhou H, Dong J, Cheng J, Dong W, Huang C, Shen Y et al. Photonic matrix multiplication lights up photonic accelerator and beyond. *Light Sci Appl* 2022; **11**: 30.

5  Mengu D, Mengu D, Mengu D, Rahman MSS, Rahman MSS, Rahman MSS et al. At the intersection of optics and deep learning: statistical inference, computing, and inverse design. *Adv Opt Photonics* 2022; **14**: 209–290.

6  Cutrona L, Leith E, Palermo C, Porcello L. Optical data processing and filtering systems. *IRE Trans Inf Theory* 1960; **6**: 386–400.

7  Hopfield JJ. Neural networks and physical systems with emergent collective computational abilities. *Proc Natl Acad Sci* 1982; **79**: 2554–2558.

8  Psaltis D, Farhat N. Optical information processing based on an associative-memory model of neural nets with thresholding and feedback. *Opt Lett* 1985; **10**: 98–100.

9  Farhat NH, Psaltis D, Prata A, Paek E. Optical implementation of the Hopfield model. *Appl Opt* 1985; **24**: 1469–1475.

10 Wagner K, Psaltis D. Multilayer optical learning networks. *Appl Opt* 1987; **26**: 5061–5076.

11 Psaltis D, Brady D, Gu X-G, Lin S. Holography in artificial neural networks. *Nature* 1990; **343**: 325.

12 Vandoorne K, Dambre J, Verstraeten D, Schrauwen B, Bienstman P. Parallel Reservoir Computing Using Optical Amplifiers. *IEEE Trans Neural Netw* 2011; **22**: 1469–1481.

13 Silva A, Monticone F, Castaldi G, Galdi V, Alù A, Engheta N. Performing Mathematical Operations with Metamaterials. *Science* 2014; **343**: 160–163.

14 Vandoorne K, Mechet P, Van Vaerenbergh T, Fiers M, Morthier G, Verstraeten D et al. Experimental demonstration of reservoir computing on a silicon photonics chip. *Nat Commun* 2014; **5**: 3541.

15 Carolan J, Harrold C, Sparrow C, Martín-López E, Russell NJ, Silverstone JW et al. Universal linear optics. *Science* 2015; **349**: 711–716.

16 Chang J, Sitzmann V, Dun X, Heidrich W, Wetzstein G. Hybrid optical-electronic convolutional neural networks with optimized diffractive optics for image classification. *Sci Rep* 2018; **8**. doi:10.1038/s41598-018-30619-y.

17 Estakhri NM, Edwards B, Engheta N. Inverse-designed metastructures that solve equations. *Science* 2019; **363**: 1333–1338.

18 Dong J, Rafayelyan M, Krzakala F, Gigan S. Optical Reservoir Computing Using Multiple Light Scattering for Chaotic Systems Prediction. *IEEE J Sel Top Quantum Electron* 2020; **26**: 1–12.

19 Teğin U, Yıldırım M, Oğuz İ, Moser C, Psaltis D. Scalable optical learning operator. *Nat Comput Sci* 2021; **1**: 542–549.

20  Shen Y, Harris NC, Skirlo S, Prabhu M, Baehr-Jones T, Hochberg M *et al.* Deep learning with coherent nanophotonic circuits. *Nat Photonics* 2017; **11**: 441–446.

21  Tait AN, Lima TF de, Zhou E, Wu AX, Nahmias MA, Shastri BJ *et al.* Neuromorphic photonic networks using silicon photonic weight banks. *Sci Rep* 2017; **7**: 7430.

22  Lin X, Rivenson Y, Yardimci NT, Veli M, Luo Y, Jarrahi M *et al.* All-optical machine learning using diffractive deep neural networks. *Science* 2018; **361**: 1004–1008.

23  Bueno J, Maktoobi S, Froehly L, Fischer I, Jacquot M, Larger L *et al.* Reinforcement learning in a large-scale photonic recurrent neural network. *Optica* 2018; **5**: 756–760.

24  Zuo Y, Li B, Zhao Y, Jiang Y, Chen Y-C, Chen P *et al.* All-optical neural network with nonlinear activation functions. *Optica* 2019; **6**: 1132–1137.

25  Hughes TW, Williamson IAD, Minkov M, Fan S. Wave physics as an analog recurrent neural network. *Sci Adv* 2019; **5**: eaay6946.

26  Feldmann J, Youngblood N, Wright CD, Bhaskaran H, Pernice WHP. All-optical spiking neurosynaptic networks with self-learning capabilities. *Nature* 2019; **569**: 208.

27  Miscuglio M, Sorger VJ. Photonic tensor cores for machine learning. *Appl Phys Rev* 2020; **7**: 031404.

28  Zhang H, Gu M, Jiang XD, Thompson J, Cai H, Paesani S *et al.* An optical neural chip for implementing complex-valued neural network. *Nat Commun* 2021; **12**: 457.

29  Feldmann J, Youngblood N, Karpov M, Gehring H, Li X, Stappers M *et al.* Parallel convolutional processing using an integrated photonic tensor core. *Nature* 2021; **589**: 52–58.

30  Xu X, Tan M, Corcoran B, Wu J, Boes A, Nguyen TG *et al.* 11 TOPS photonic convolutional accelerator for optical neural networks. *Nature* 2021; **589**: 44–51.

31  Wright LG, Onodera T, Stein MM, Wang T, Schachter DT, Hu Z *et al.* Deep physical neural networks trained with backpropagation. *Nature* 2022; **601**: 549–555.

32  Ashtiani F, Geers AJ, Aflatouni F. An on-chip photonic deep neural network for image classification. *Nature* 2022; **606**: 501–506.

33  Liu D, Tan Y, Khoram E, Yu Z. Training Deep Neural Networks for the Inverse Design of Nanophotonic Structures. *ACS Photonics* 2018; **5**: 1365–1369.

34  Ma W, Cheng F, Liu Y. Deep-Learning-Enabled On-Demand Design of Chiral Metamaterials. *ACS Nano* 2018; **12**: 6326–6334.

35  Peurifoy J, Shen Y, Jing L, Yang Y, Cano-Renteria F, DeLacy BG *et al.* Nanophotonic particle simulation and inverse design using artificial neural networks. *Sci Adv* 2018; **4**: eaar4206.

36  Malkiel I, Mrejen M, Nagler A, Arieli U, Wolf L, Suchowski H. Plasmonic nanostructure design and characterization via Deep Learning. *Light Sci Appl* 2018; **7**: 60.

37  Liu Z, Zhu D, Rodrigues SP, Lee K-T, Cai W. Generative Model for the Inverse Design of Metasurfaces. *Nano Lett* 2018; **18**: 6570–6576.

38  So S, Rho J. Designing nanophotonic structures using conditional deep convolutional generative adversarial networks. *Nanophotonics* 2019; **8**: 1255–1261.

39 Ma W, Cheng F, Xu Y, Wen Q, Liu Y. Probabilistic Representation and Inverse Design of Metamaterials Based on a Deep Generative Model with Semi-Supervised Learning Strategy. *Adv Mater* 2019; **31**: 1901111.

40 An S, Fowler C, Zheng B, Shalaginov MY, Tang H, Li H *et al.* A Deep Learning Approach for Objective-Driven All-Dielectric Metasurface Design. *ACS Photonics* 2019; **6**: 3196–3207.

41 Jiang J, Sell D, Hoyer S, Hickey J, Yang J, Fan JA. Free-Form Diffractive Metagrating Design Based on Generative Adversarial Networks. *ACS Nano* 2019; **13**: 8872–8878.

42 Qian C, Zheng B, Shen Y, Jing L, Li E, Shen L *et al.* Deep-learning-enabled self-adaptive microwave cloak without human intervention. *Nat Photonics* 2020; **14**: 383–390.

43 Liu Z, Zhu D, Lee K-T, Kim AS, Raju L, Cai W. Compounding Meta-Atoms into Metamolecules with Hybrid Artificial Intelligence Techniques. *Adv Mater* 2020; **32**: 1904790.

44 Ren H, Shao W, Li Y, Salim F, Gu M. Three-dimensional vectorial holography based on machine learning inverse design. *Sci Adv* 2020; **6**: eaaz4261.

45 Zuo C, Chen Q. Exploiting optical degrees of freedom for information multiplexing in diffractive neural networks. *Light Sci Appl* 2022; **11**: 208.

46 Mengu D, Luo Y, Rivenson Y, Ozcan A. Analysis of Diffractive Optical Neural Networks and Their Integration With Electronic Neural Networks. *IEEE J Sel Top Quantum Electron* 2020; **26**: 1–14.

47 Li J, Mengu D, Luo Y, Rivenson Y, Ozcan A. Class-specific differential detection in diffractive optical neural networks improves inference accuracy. *Adv Photonics* 2019; **1**: 046001.

48 Yan T, Wu J, Zhou T, Xie H, Xu F, Fan J *et al.* Fourier-space Diffractive Deep Neural Network. *Phys Rev Lett* 2019; **123**: 023901.

49 Mengu D, Rivenson Y, Ozcan A. Scale-, Shift-, and Rotation-Invariant Diffractive Optical Networks. *ACS Photonics* 2020. doi:10.1021/acsphotonics.0c01583.

50 Mengu D, Zhao Y, Yardimci NT, Rivenson Y, Jarrahi M, Ozcan A. Misalignment resilient diffractive optical networks. *Nanophotonics* 2020; **9**: 4207–4219.

51 Rahman MSS, Li J, Mengu D, Rivenson Y, Ozcan A. Ensemble learning of diffractive optical networks. *Light Sci Appl* 2021; **10**: 14.

52 Li J, Mengu D, Yardimci NT, Luo Y, Li X, Veli M *et al.* Spectrally encoded single-pixel machine vision using diffractive networks. *Sci Adv* 2021; **7**: eabd7690.

53 Kulce O, Mengu D, Rivenson Y, Ozcan A. All-optical information-processing capacity of diffractive surfaces. *Light Sci Appl* 2021; **10**: 25.

54 Zhou T, Lin X, Wu J, Chen Y, Xie H, Li Y *et al.* Large-scale neuromorphic optoelectronic computing with a reconfigurable diffractive processing unit. *Nat Photonics* 2021; **15**: 367–373.

55 Chen H, Feng J, Jiang M, Wang Y, Lin J, Tan J *et al.* Diffractive Deep Neural Networks at Visible Wavelengths. *Engineering* 2021; **7**: 1483–1491.

56 Liu C, Ma Q, Luo ZJ, Hong QR, Xiao Q, Zhang HC *et al.* A programmable diffractive deep neural network based on a digital-coding metasurface array. *Nat Electron* 2022; : 1–10.

57 Mengu D, Veli M, Rivenson Y, Ozcan A. Classification and reconstruction of spatially overlapping phase images using diffractive optical networks. *Sci Rep* 2022; **12**: 8446.

58 Luo Y, Zhao Y, Li J, Çetintaş E, Rivenson Y, Jarrahi M *et al.* Computational imaging without a computer: seeing through random diffusers at the speed of light. *eLight* 2022; **2**: 4.

59 Mengu D, Zhao Y, Tabassum A, Jarrahi M, Ozcan A. Diffractive Interconnects: All-Optical Permutation Operation Using Diffractive Networks. 2022. doi:10.48550/arXiv.2206.10152.

60 Mengu D, Ozcan A. All-Optical Phase Recovery: Diffractive Computing for Quantitative Phase Imaging. *Adv Opt Mater* 2022; **10**: 2200281.

61 Bai B, Luo Y, Gan T, Hu J, Li Y, Zhao Y *et al.* To image, or not to image: Class-specific diffractive cameras with all-optical erasure of undesired objects. ; : 31.

62 Isil C, Mengu D, Zhao Y, Tabassum A, Li J, Luo Y *et al.* Super-resolution image display using diffractive decoders. 2022. doi:10.48550/arXiv.2206.07281.

63 Qian C, Lin X, Lin X, Xu J, Sun Y, Li E *et al.* Performing optical logic operations by a diffractive neural network. *Light Sci Appl* 2020; **9**: 1–7.

64 Wang P, Xiong W, Huang Z, He Y, Xie Z, Liu J *et al.* Orbital angular momentum mode logical operation using optical diffractive neural network. *Photonics Res* 2021; **9**: 2116–2124.

65 Luo Y, Mengu D, Ozcan A. Cascadable all-optical NAND gates using diffractive networks. *Sci Rep* 2022; **12**: 7121.

66 Luo Y, Mengu D, Yardimci NT, Rivenson Y, Veli M, Jarrahi M *et al.* Design of task-specific optical systems using broadband diffractive neural networks. *Light Sci Appl* 2019; **8**: 1–14.

67 Veli M, Mengu D, Yardimci NT, Luo Y, Li J, Rivenson Y *et al.* Terahertz pulse shaping using diffractive surfaces. *Nat Commun* 2021; **12**: 37.

68 Huang Z, Wang P, Liu J, Xiong W, He Y, Xiao J *et al.* All-Optical Signal Processing of Vortex Beams with Diffractive Deep Neural Networks. *Phys Rev Appl* 2021; **15**: 014037.

69 Kulce O, Mengu D, Rivenson Y, Ozcan A. All-optical synthesis of an arbitrary linear transformation using diffractive surfaces. *Light Sci Appl* 2021; **10**: 196.

70 Li J, Hung Y-C, Kulce O, Mengu D, Ozcan A. Polarization multiplexed diffractive computing: all-optical implementation of a group of linear transformations through a polarization-encoded diffractive network. *Light Sci Appl* 2022; **11**: 153.

71 TSL-570 | SANTEC CORPORATION - The Photonics Pioneer. https://www.santec.com/en/products/instruments/tunablelaser/TSL-570/ (accessed 1 Aug2022).

72 MEMS-VCSEL Swept-Wavelength Laser Sources. https://www.thorlabs.com/newgrouppage9.cfm?objectgroup_id=12057 (accessed 1 Aug2022).

73 Kingma DP, Ba J. Adam: A Method for Stochastic Optimization. *ArXiv14126980 Cs* 2014.http://arxiv.org/abs/1412.6980 (accessed 16 Apr2019).

74 Loshchilov I, Hutter F. Decoupled Weight Decay Regularization. In: *International Conference on Learning Representations 2019*. 2019, p 18.

# Figures

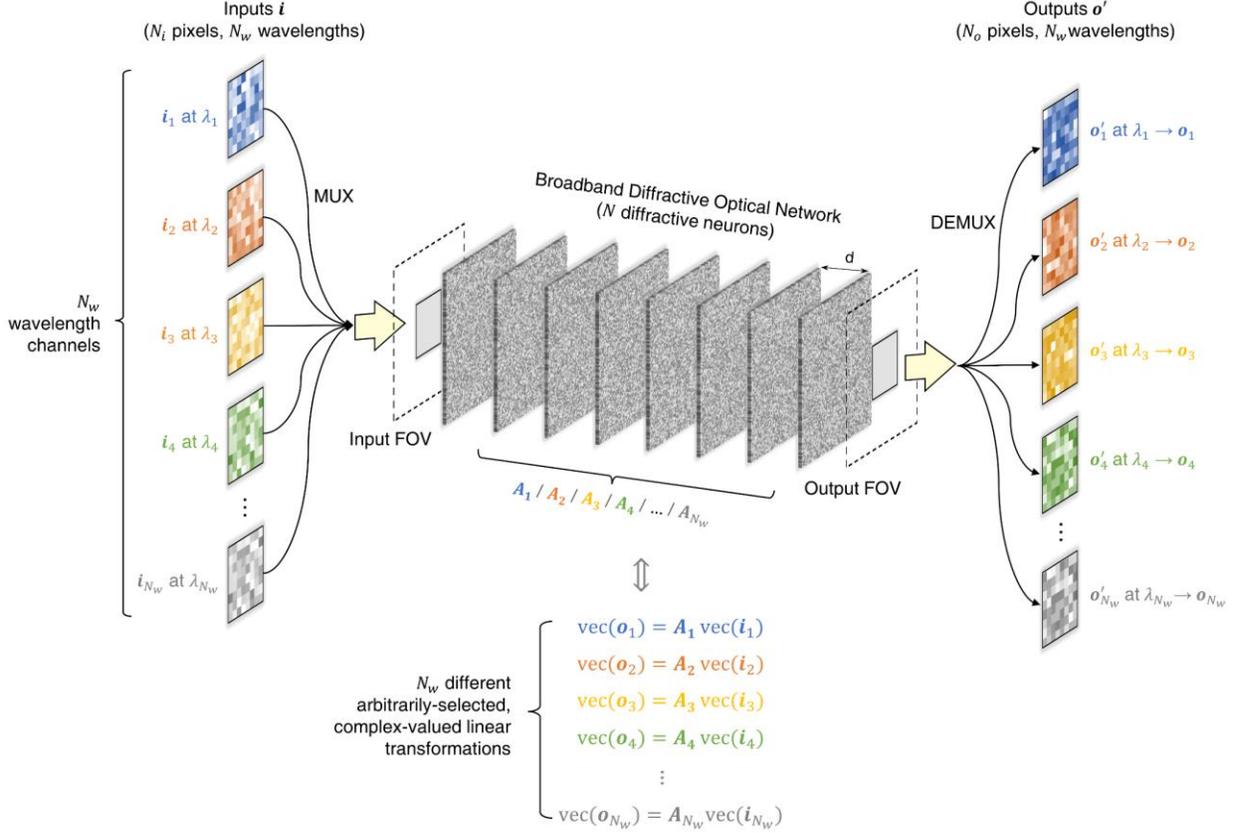

**Figure 1. Schematic of massively parallel, wavelength-multiplexed diffractive optical computing. A**, Optical layout of the wavelength multiplexed diffractive neural network, where the diffractive layers are jointly trained to perform $N_w$ different arbitrarily-selected, complex-valued linear transformations between the input field $i$ and the output field $o'$ using wavelength-multiplexing. The optical fields at the input FOV, $i_1, i_2, \ldots, i_{N_w}$, are encoded at a predetermined set of distinct wavelengths $\lambda_1, \lambda_2, \ldots, \lambda_{N_w}$, respectively, using a wavelength multiplexing ("MUX") scheme. At the output FOV of the broadband diffractive network, wavelength demultiplexing ("DEMUX") is performed to extract the diffractive output fields $o'_1, o'_2, \ldots, o'_{N_w}$ at the corresponding wavelengths $\lambda_1, \lambda_2, \ldots, \lambda_{N_w}$, respectively, which represent the all-optical estimates of the target output fields $o_1, o_2, \ldots, o_{N_w}$, corresponding to the target linear transformations ($A_1, A_2, \ldots, A_{N_w}$). Through this diffractive architecture, $N_w$ different arbitrarily-selected complex-valued linear transformations can be all-optically performed at distinct wavelengths, running in parallel channels of the broadband diffractive processor.

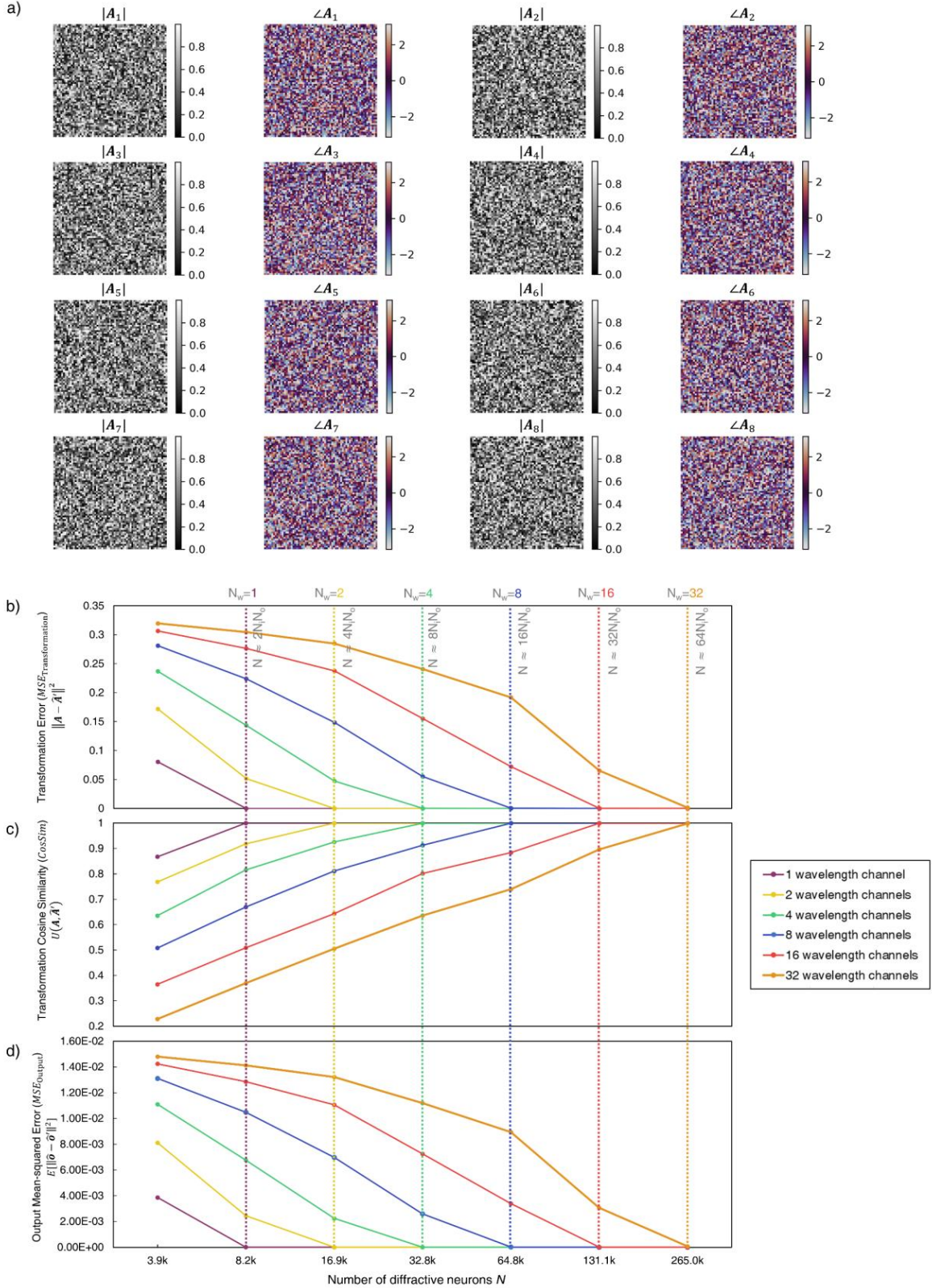

**Figure 2. All-optical transformation performances of broadband diffractive networks using different number of wavelength channels. a**, As examples, we show the amplitude and phase of

the first 8 matrices in $\{A_1, A_2, \ldots, A_{32}\}$ that were randomly generated, serving as the ground truth (target) for the diffractive all-optical transformations. See Supplementary Fig. S1 for the cosine similarity values calculated between any two combinations of these 32 target linear transformation matrices. **b**, The mean values of the normalized mean-squared error between the ground truth transformation matrices ($A_w$) and the corresponding all-optical transforms ($A'_w$) across different wavelength channels are reported as a function of the number of diffractive neurons, $N$. The results of the diffractive networks using different number of wavelength channels ($N_w$) are encoded with different colors, and the space between the simulation data points is linearly interpolated. $N_w \in \{1, 2, 4, 8, 16 \text{ and } 32\}$, $N \in \{3.9k, 8.2k, 16.9k, 32.8k, 64.8k, 131.1k, 265.0k\}$ and $N_i = N_o = 8^2$. **c**, Same as (b) but the cosine similarity values between the all-optical transforms and their ground truth are reported. **d**, Same as (b) but the mean-squared error values between the diffractive network output fields and the ground truth output fields are reported.

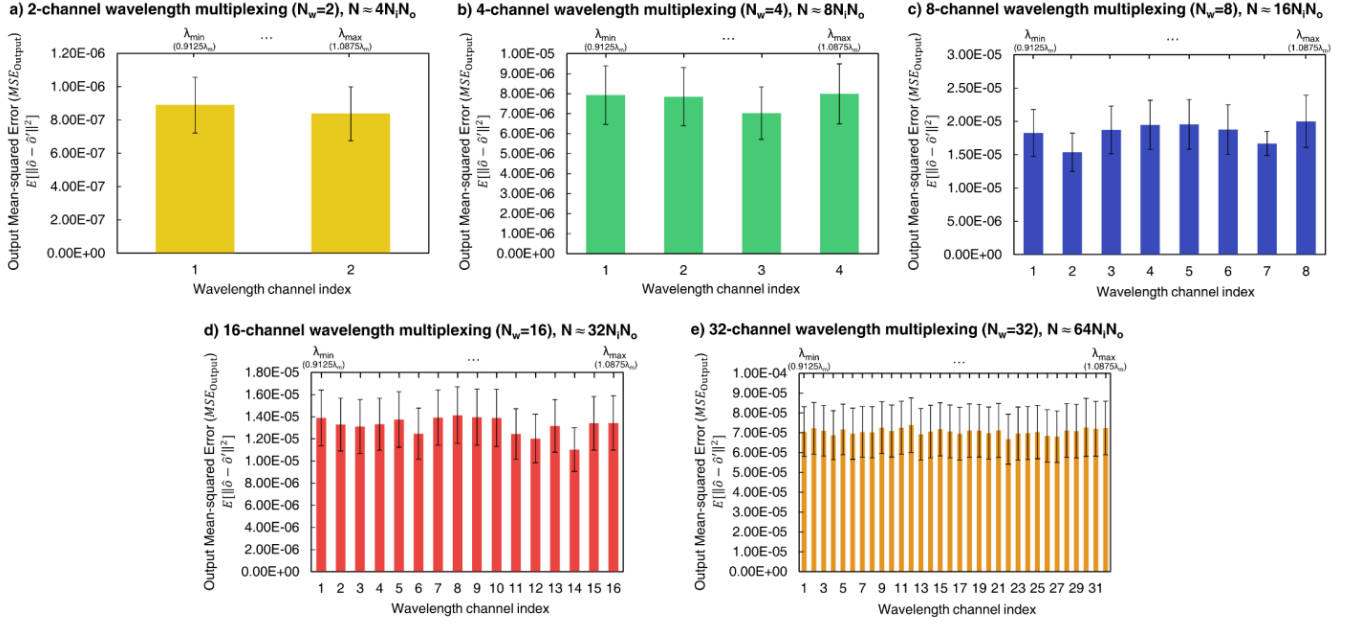

**Figure 3. All-optical transformation performances of the individual wavelength channels in broadband diffractive network designs with $N \approx 2N_wN_iN_o$ and $N_i = N_o = 8^2$.** The output field errors ($MSE_{\text{Output}}$) for the all-optical linear transforms achieved by the wavelength-multiplexed diffractive network models with $N_w = 2$ (**a**), 4 (**b**), 8 (**c**), 16 (**d**) and 32 (**e**) are shown. The standard deviations (error bars) of these metrics are calculated across the entire testing dataset.

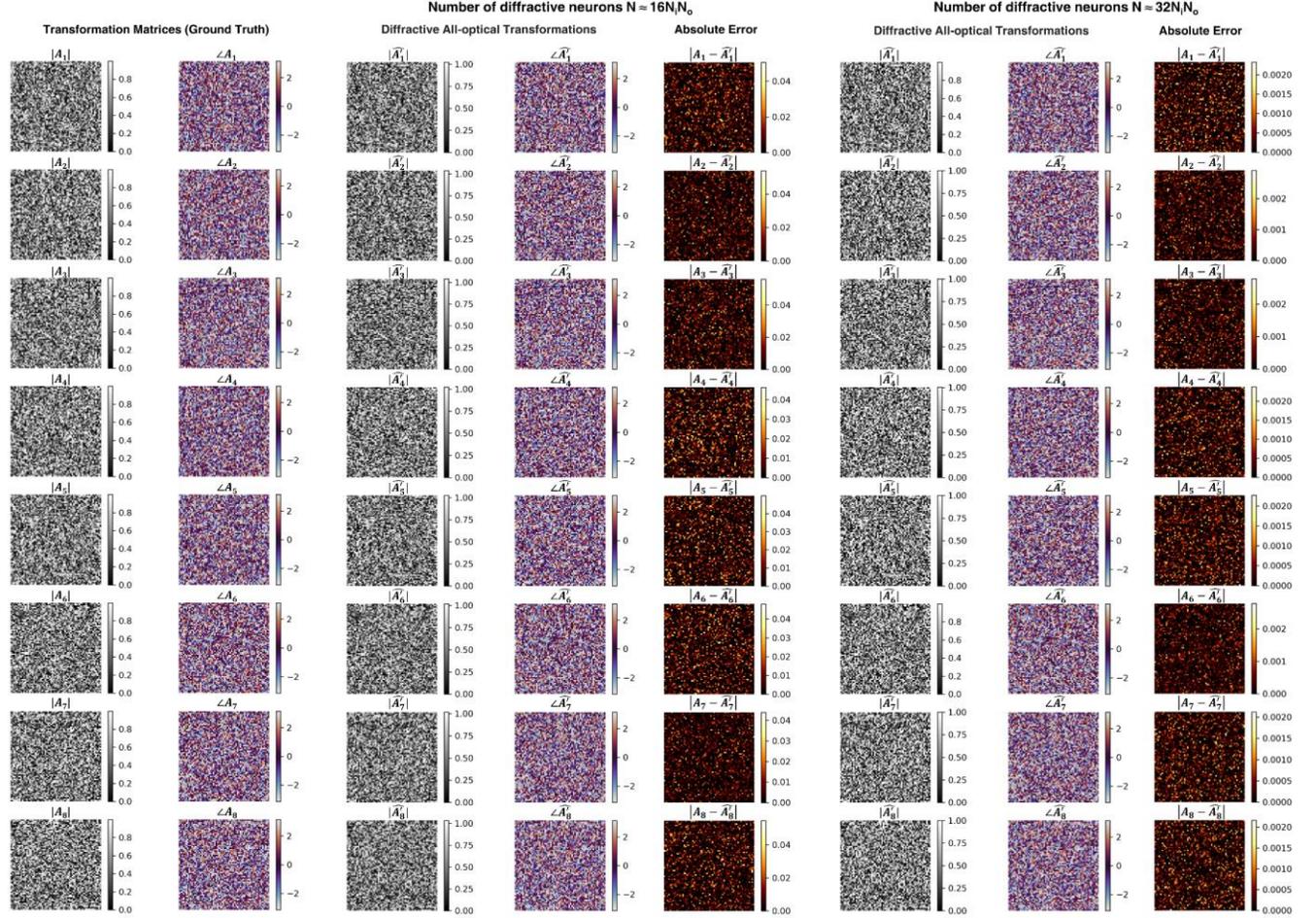

**Figure 4. All-optical transformation matrices estimated by two different wavelength-multiplexed broadband diffractive networks with $N_w = 8$ and $N_i = N_o = 8^2$.** The first broadband diffractive network has $N \approx 2N_w N_i N_o = 16 N_i N_o = 64.8$k trainable diffractive neurons. The second broadband diffractive network has $N \approx 4N_w N_i N_o = 32 N_i N_o = 131.1$k trainable diffractive neurons. The absolute differences between these all-optical transformation matrices and the corresponding ground truth (target) matrices are also shown in each case. $N = 131.1k$ diffractive design achieves a much smaller and negligible absolute error due to the increased degrees of freedom.

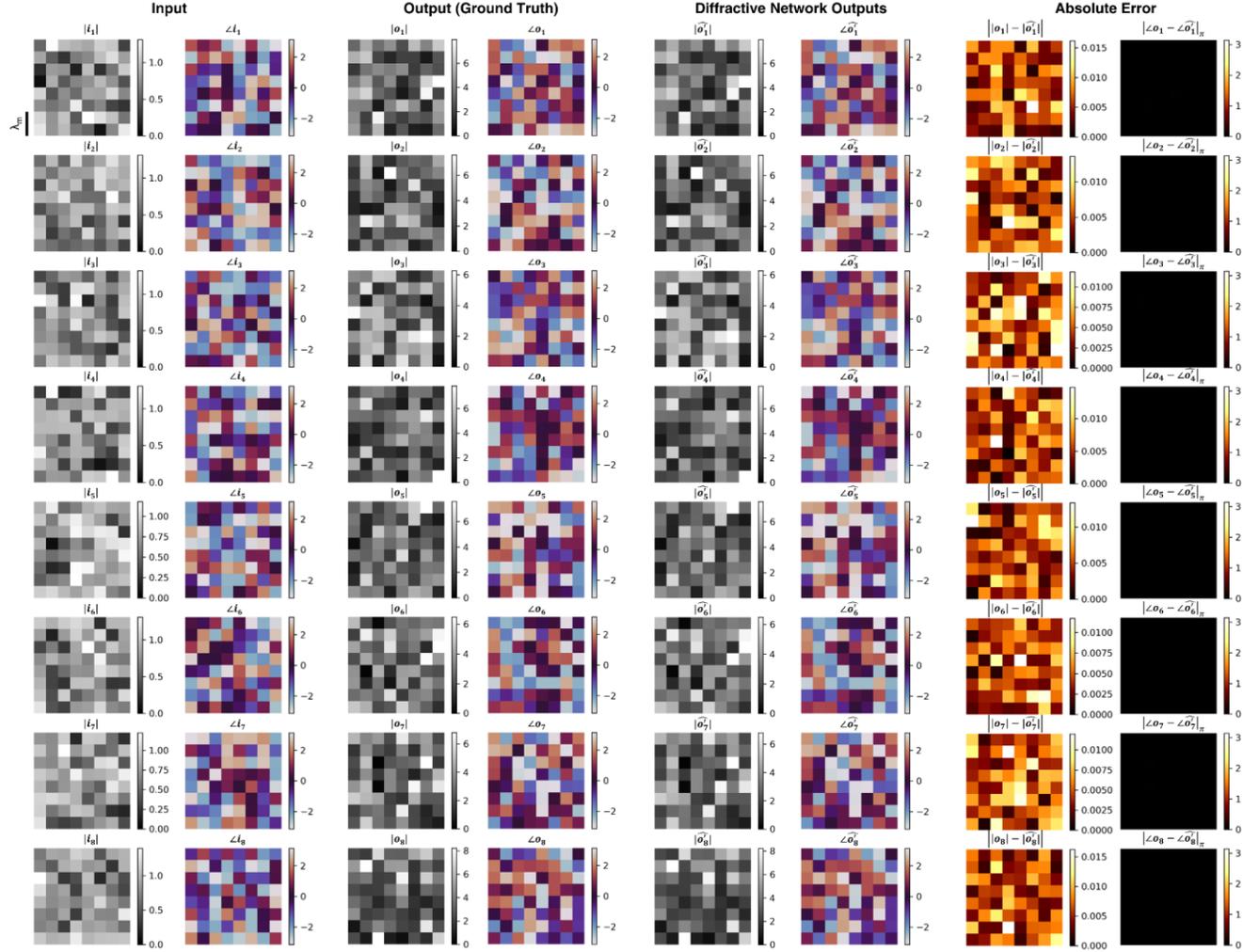

**Figure 5. Examples of the input/output complex fields for the ground truth (target) transformations along with the all-optical output fields resulting from the 8-channel wavelength-multiplexed diffractive design using $N \approx 4N_w N_i N_o = 32 N_i N_o = 131.1k$.** Absolute errors between the ground truth output fields and the all-optical diffractive network output fields are negligible. Note that $\left|\angle o - \angle \widehat{o'}\right|_\pi$ indicates the wrapped phase difference between the ground truth output field $o$ and the normalized diffractive network output field $\widehat{o'}$.

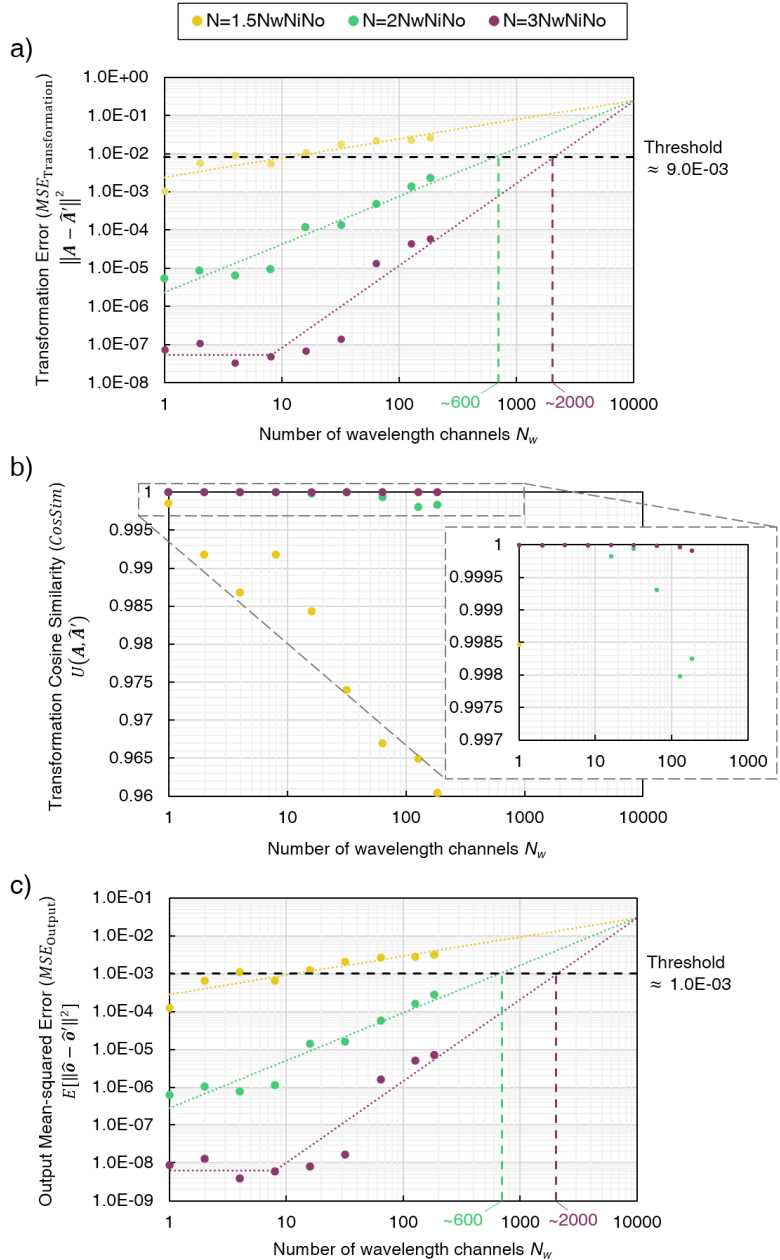

**Figure 6. Exploration of the limits of the number of wavelength channels ($N_w$) that can be implemented in a broadband diffractive network. a**, The mean values of the normalized mean-squared error between the ground truth transformation matrices ($A_w$) and the all-optical transforms ($A'_w$) across different wavelength channels are reported as a function of $N_w \in \{1, 2, 4, 8, 16, 32, 64, 128, 184\}$. The results of the broadband diffractive networks using different numbers of diffractive neurons ($N$) are presented with different colors: $N \in \{1.5 N_w N_i N_o, \ 2 N_w N_i N_o, \ 3 N_w N_i N_o\}$. Dotted lines are fitted based on the data points whose diffractive networks share the same $N$. **b**, Same as (a) but the cosine similarity values between the all-optical transforms and their ground truth are reported. **c**, Same as (a) but the mean-squared error values between the diffractive network output fields and the ground truth output fields are reported. $N_i = N_o = 5^2$.

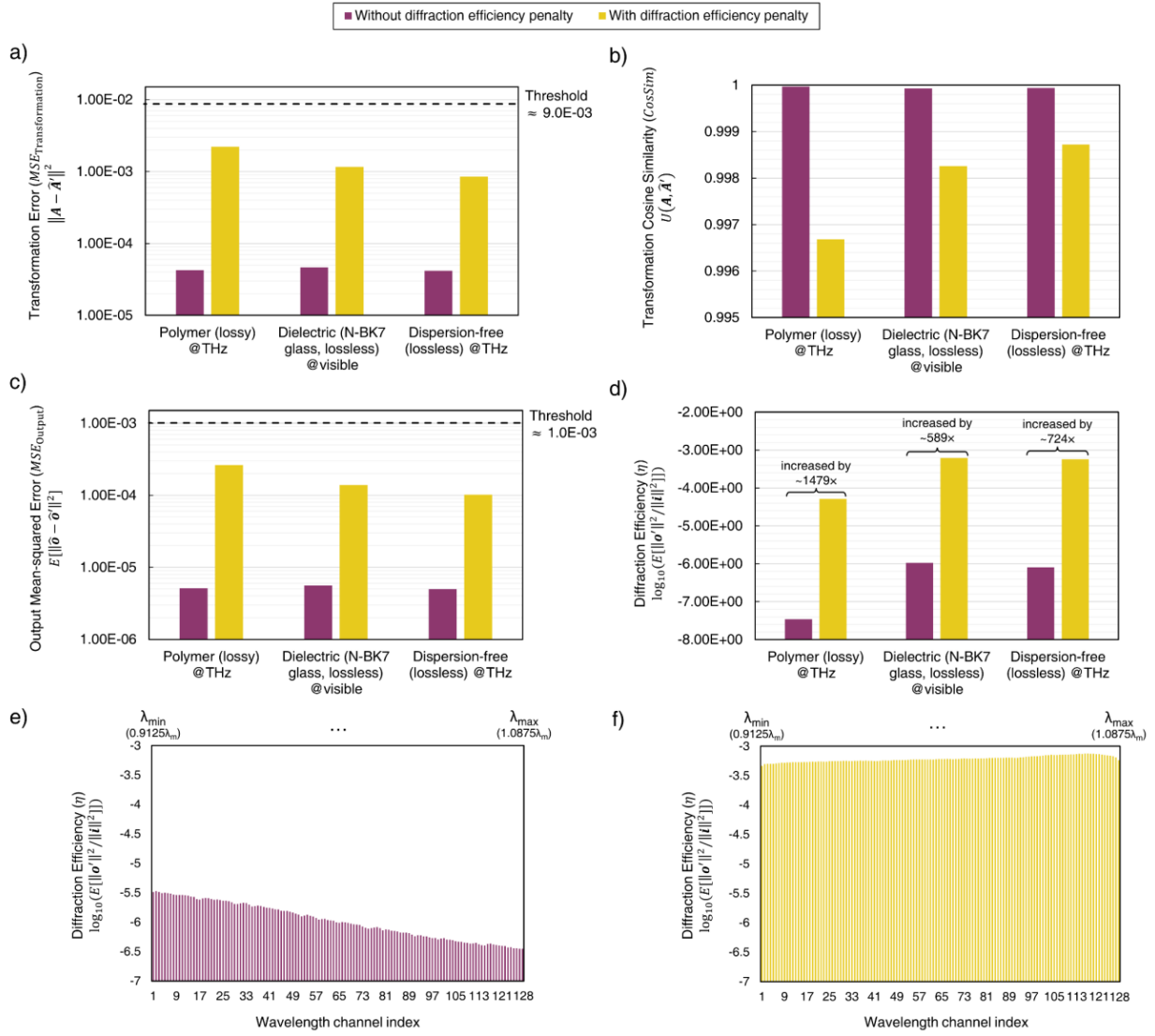

**Figure 7. The impact of material dispersion and losses on the all-optical transformation performance of wavelength-multiplexed broadband diffractive networks. a**, The mean values of the normalized mean-squared error between the ground truth transformation matrices ($A_w$) and the all-optical transforms ($A'_w$) across different wavelength channels are reported as a function of the material of the diffractive layers. The results of the diffractive networks trained with and without diffraction efficiency penalty are presented in yellow and purple colors, respectively. $N_w = 128$, $N = 3N_w N_i N_o$ and $N_i = N_o = 5^2$. **b**, Same as (a) but the cosine similarity values between the all-optical transforms and their ground truth are reported. **c**, Same as (a) but the mean-squared error values between the diffractive network output fields and the ground truth fields are reported. **d**, The mean diffraction efficiencies of the presented diffractive models across all the wavelength channels. **e**, Diffraction efficiency of the individual wavelength channels for the broadband diffractive network model presented in (a)-(d) that uses the dielectric material without the diffraction efficiency-related penalty term in its loss function. **f**, Same as (e), but the diffractive network was trained using a loss function *with* the diffraction efficiency-related penalty term.

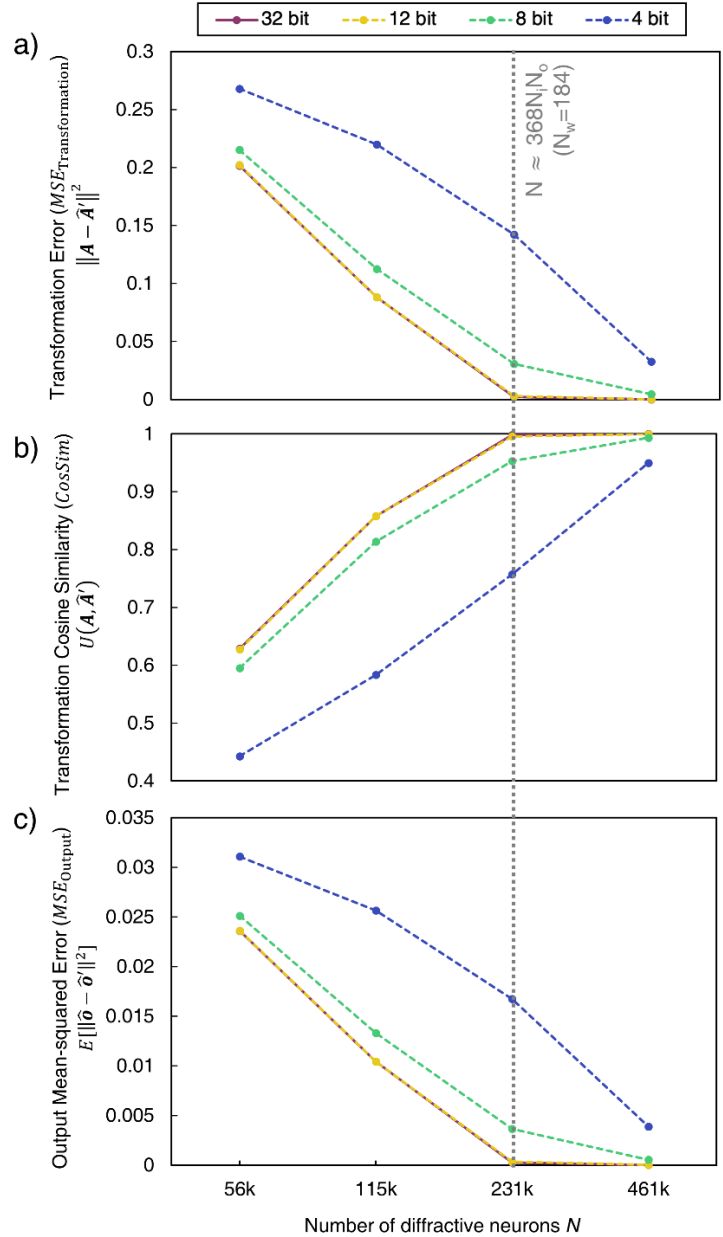

**Figure 8. All-optical transformation performance of broadband diffractive network designs with $N_w = 184$, reported as a function of $N$ and the bit depth of the diffractive neurons. a,** The mean values of normalized mean-squared error between the ground truth transformation matrices ($A_w$) and the all-optical transforms ($A'_w$) across different wavelength channels are reported as a function of $N$. The results of the diffractive networks using different bit depths of the diffractive neurons, including 4, 8, 12 and 32, are encoded with different colors, and the space between the data points is linearly interpolated. $N \in \{0.5 N_w N_i N_o = 56\text{k}, N_w N_i N_o = 115\text{k}, 2N_w N_i N_o = 231\text{k}, 4N_w N_i N_o = 461\text{k}\}$, and $N_i = N_o = 5^2$. **b,** Same as (a) but the cosine similarity values between the all-optical transforms and their ground truth are reported. **c,** Same as (a) but the mean-squared error values between the diffractive network output fields and the ground truth output fields are reported.

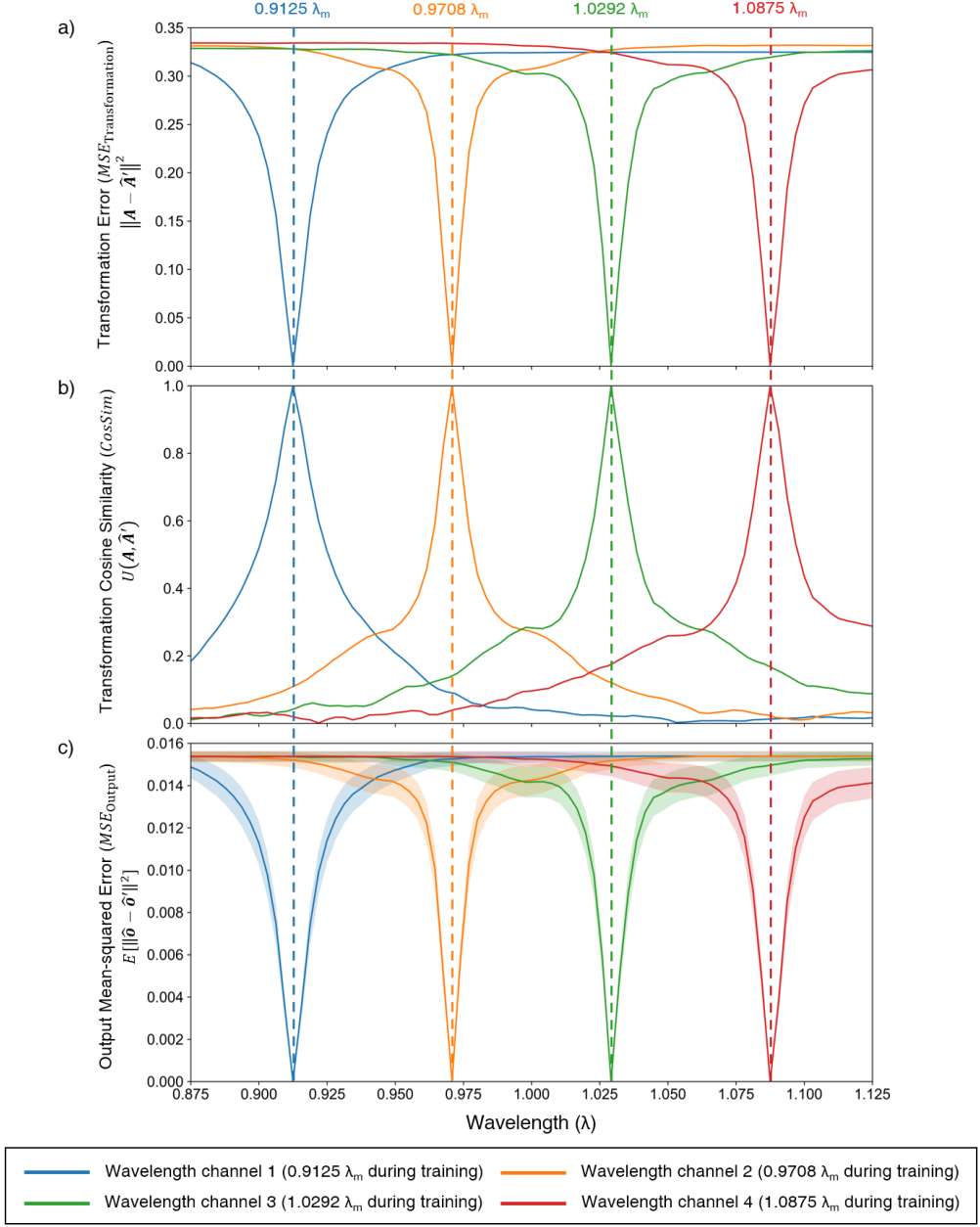

**Figure 9. The impact of the encoding wavelength error on the all-optical linear transformation performance of a wavelength-multiplexed broadband diffractive network;** $N_w = 4$, $N \approx 2N_w N_i N_o = 8N_i N_o$ and $N_i = N_o = 8^2$. **a,** The normalized mean-squared error values between the ground truth transformation matrices ($A_w$) and the all-optical transforms ($A'_w$) for the four different wavelength channels are reported as a function of the wavelengths used during the testing. The results of the different wavelength channels are shown with different colors, and the space between the simulation data points is linearly interpolated. **b,** Same as (a) but the cosine similarity values between the all-optical transforms and their ground truth are reported. **c,** Same as (a) but the mean-squared error values between the diffractive network output fields and the ground truth output fields are reported. The shaded areas indicate the standard deviation values calculated based on all the samples in the testing dataset.